\documentclass[10pt,twocolumn,letterpaper]{article}

\usepackage{iccv}
\usepackage{times}
\usepackage{epsfig}
\usepackage{graphicx}
\usepackage{amsmath}
\usepackage{amssymb}

\usepackage[utf8]{inputenc} 
\usepackage{array} 
\usepackage{multirow} 

\graphicspath{{images/}} 
\DeclareGraphicsExtensions{.pdf,.jpeg,.jpg,.JPG,.png} 

\newcommand{\diff}[1]{\mathop{}\!\mathrm{d}{#1}}

\newcommand{\iali}[1]{\begin{align}#1\end{align}}
\newcommand{\ialid}[1]{\begin{aligned}#1\end{aligned}}
\newcommand{\ieqn}[1]{\begin{equation}#1\end{equation}}

\newcommand{\imtx}[1]{\begin{bmatrix}#1\end{bmatrix}}

\newcommand{\bfg}{\begin{figure}}
\newcommand{\efg}{\end{figure}}
\newcommand{\bsl}{\begin{slide}}
\newcommand{\esl}{\end{slide}}
\newcommand{\bthm}{\begin{theorem}}
\newcommand{\ethm}{\end{theorem}}
\newcommand{\bcor}{\begin{corollary}}
\newcommand{\ecor}{\end{corollary}}
\newcommand{\blem}{\begin{lemma}}
\newcommand{\elem}{\end{lemma}}
\newcommand{\bprop}{\begin{proposition}}
\newcommand{\eprop}{\end{proposition}}
\newcommand{\basm}{\begin{assumption}}
\newcommand{\easm}{\end{assumption}}
\newcommand{\baxm}{\begin{axiom}}
\newcommand{\eaxm}{\end{axiom}}
\newcommand{\bdfn}{\begin{definition}}
\newcommand{\edfn}{\end{definition}}
\newcommand{\brmk}{\begin{remark}}
\newcommand{\ermk}{\end{remark}}
\newcommand{\balg}{\begin{algorithm}}
\newcommand{\ealg}{\end{algorithm}}
\newcommand{\bntn}{\begin{notation}}
\newcommand{\entn}{\end{notation}}
\newcommand{\bexm}{\begin{example}}
\newcommand{\eexm}{\end{example}}
\newcommand{\bpf}{\begin{proof}}
\newcommand{\epf}{\end{proof}}

\newcommand{\st}{\text{s.t.~}}

\newcommand{\bR}{\mathbb{R}}

\newcommand{\bS}{\mathbb{S}}

\newcommand{\fp}{\mathbf{p}}

\newcommand{\fx}{\mathbf{x}}

\newcommand{\fl}{\mathbf{l}}

\newcommand{\fn}{\mathbf{n}}

\newcommand{\fh}{\mathbf{h}}

\renewcommand{\hat}{\widehat}
\renewcommand{\tilde}{\widetilde}

\newcommand{\bomega}{{\boldsymbol{\omega}}}
\newcommand{\tn}{\tilde{\fn}}
\newcommand{\tz}{\hat{z}}

\usepackage[pagebackref=true,breaklinks=true,letterpaper=true,colorlinks,bookmarks=false]{hyperref}

\iccvfinalcopy 

\def\iccvPaperID{1335} 

\ificcvfinal  \fi

\begin{document}

\title{Variational Uncalibrated Photometric Stereo under General Lighting}

\author{Bjoern Haefner\thanks{Authors contributed equally.} \textsuperscript{$~$,$~$1}   \quad
	Zhenzhang Ye\footnotemark[1] \textsuperscript{$~$,$~$1} \quad
	Maolin Gao\textsuperscript{2} \quad
	Tao Wu\textsuperscript{1} \quad
	Yvain Qu\'eau\textsuperscript{3}\quad 
	Daniel Cremers\textsuperscript{1}\\
	\textsuperscript{1}Technical University of Munich \quad \textsuperscript{2}Artisense \quad \textsuperscript{3}GREYC, UMR CNRS 6072\\
	{\tt\small \{bjoern.haefner, zz.ye, tao.wu, cremers\}@tum.de
		{maolin@artisense.ai}
		{yvain.queau@ensicaen.fr}}
}

\maketitle
\ificcvfinal \thispagestyle{empty} \fi

\begin{abstract}
 Photometric stereo (PS) techniques nowadays remain constrained to an ideal laboratory setup where modeling and calibration of lighting is amenable. To eliminate such restrictions, we propose an efficient principled variational approach to uncalibrated PS under general illumination. To this end, the Lambertian reflectance model is approximated through a spherical harmonic expansion, which preserves the spatial invariance of the lighting.
The joint recovery of shape, reflectance and illumination is then formulated as a single variational problem. There the shape estimation is carried out directly in terms of the underlying perspective depth map, thus implicitly ensuring integrability and bypassing the need for a subsequent normal integration. To tackle the resulting nonconvex problem numerically, we undertake a two-phase procedure to initialize a balloon-like perspective depth map, followed by a ``lagged'' block coordinate descent scheme. The experiments validate efficiency and robustness of this approach. Across a variety of evaluations, we are able to reduce the mean angular error consistently by a factor of $2$--$3$ compared to the state-of-the-art.

\end{abstract}

\section{Introduction}

Photometric stereo techniques aim at acquiring both the shape and the reflectance of a scene. To this end, multiple images are acquired under the same viewing angle but varying lighting, and a physics-based image formation model is inverted. However, the classic way to solve this inverse problem requires lighting to be highly controlled, which restricts practical applications to laboratory setups where careful calibration of lighting must be carried out.

\begin{figure}[!ht]
  \centering
  \newcommand{\mywidth}{0.115\textwidth}
  \newcommand{\mywidthy}{0.1875\textwidth}
  \newcolumntype{C}{ >{\centering\arraybackslash} m{0.02\textwidth} }
  \newcolumntype{X}{ >{\centering\arraybackslash} m{\mywidth} }
  \newcolumntype{Y}{ >{\centering\arraybackslash} m{\mywidthy} }
  \newcommand{\tabelt}[1]{\hfil\hbox to 0pt{\hss #1 \hss}\hfil}
  \setlength\tabcolsep{1pt} 
  \def\arraystretch{1} 
  \begin{tabular}{CXXCX}
    &$I^1$&$I^2$&\dots&$I^M$\\
    \rotatebox{90}{Input}&
    \includegraphics[width=\mywidth]{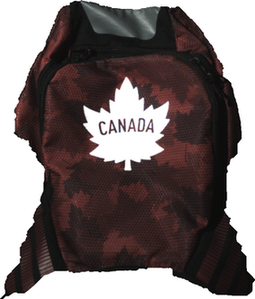}&
    \includegraphics[width=\mywidth]{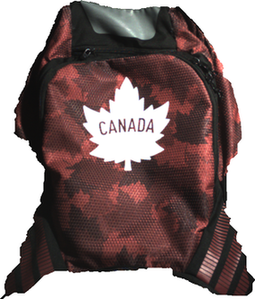}&
    \dots&
    \includegraphics[width=\mywidth]{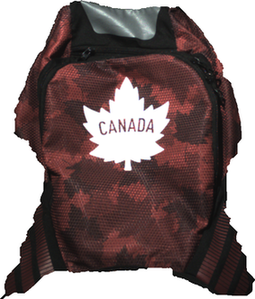}
    \end{tabular}
    \begin{tabular}{CYY}
    \rotatebox{90}{Output}&
    \includegraphics[width=\mywidth]{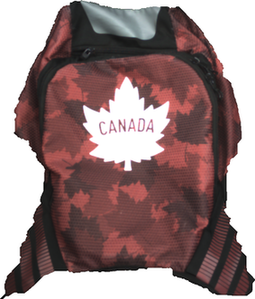}&
    \includegraphics[width=\mywidth]{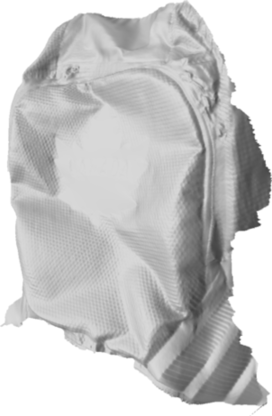}\\[-.75em]
    &Reflectance & Shape
  \end{tabular}
  \caption{We present an efficient variational scheme to solve uncalibrated photometric stereo under general lighting. Given a set of input RGB images captured from the same viewing angle but under unknown, varying general illumination (top, $M=20$ images were acquired in an office under daylight, while freely moving a hand-held LED light source), fine-detailed reflectance and shape (bottom, we show the estimated albedo and perspective depth maps) are recovered by an end-to-end variational approach.}
\label{fig:teaser}
\end{figure}
The objective of this research work is to simplify the overall photometric stereo pipeline, by providing an efficient solution to uncalibrated photometric stereo under general lighting, as illustrated in Figure~\ref{fig:teaser} (the code is released\footnote{\url{https://github.com/zhenzhangye/general_ups}}). In comparison with existing efforts in the same direction, the proposed one has the following advantages:  
\begin{itemize}
\item The joint estimation of shape, reflectance and general lighting is formulated as an end-to-end, mathematically transparent variational problem;
\item A real 3D-surface represented as a depth map is recovered, rather than possibly non-integrable normals;
\item It is robust, due to the use of Cauchy's robust M-estimator and Huber-TV albedo regularization;
\item It is computationally efficient, thanks to a tailored lagged block coordinate descent scheme initialized using a simple balloon-like shape.
\end{itemize}

After reviewing related works in Section~\ref{sec:2}, we discuss in Section~\ref{sec:ifsha} the image formation model considered in this work. It can be inverted using the variational approach in Section~\ref{sec:vups}. A dedicated numerical solution is then introduced in Section~\ref{sec:5} and empirically evaluated in Section~\ref{sec:6}. Section~\ref{sec:7} eventually draws the conclusion of this research.

\section{Related Work}
\label{sec:2}

3D-models of scenes are essential in many applications such as visual inspection~\cite{Farooq2005} or computer-aided surgery using augmented reality~\cite{Collins2012}. A 3D-model consists of geometric (position, orientation, etc.) and photometric (color, texture, etc.) properties. 
Given a set of photographies, the aim of 3D scanning is to invert the image formation process in order to recover these geometric and photometric properties of the observed scene. This notion thus includes both those of 3D-reconstruction (geometry) and of reflectance estimation (photometry). 

Many approaches to the problem of 3D-reconstruction from photographies have been studied, and they are grouped under the generic naming ``shape-from-X'', where X stands for the clue which is being used (shadows~\cite{Shafer1983}, contours~\cite{Brady1984}, texture~\cite{Witkin1981}, template~\cite{Bartoli2015}, structured light~\cite{Geng2011}, motion~\cite{Moons2008}, focus~\cite{Nayar1994}, silhouettes~\cite{Hernandez2004}, etc.). Geometric shape-from-X techniques are based on the identification and analysis of feature point or areas in the image. In contrast, photometric techniques build upon the analysis of the quantity of light received by each photosite of the camera's sensor. Among photometric techniques, \emph{shape-from-shading} is probably the most famous one. This technique, developed in the 70s by Horn \textit{et al.}~\cite{Horn1970}, consists in 3D-reconstruction from a single image of a shaded scene. It is a classic ill-posed inverse problem whose numerical solving usually requires the surface's reflectance to be known~\cite{Durou2008}. In order both to limit the ambiguities of shape-from-shading and to allow for automatic reflectance estimation, it has been suggested to consider not just one image of the scene, but several ones acquired from the same viewing angle but under varying lighting. This variant, which was introduced in the late 70s by Woodham~\cite{Woodham1978}, is known as \emph{photometric stereo}. 

Among the various shape-from-X techniques mentioned above, photometric stereo is the only 3D-scanning technique i.e., the only one which is able to achieve both 3D-reconstruction and reflectance estimation. However, early photometric approaches strongly rely on the control of lighting. The latter is usually assumed for simplicity to be directional, although the case of nearby point light sources has recently regained some attention~\cite{logothetis2017semi,mecca2014near}. More importantly, lighting is assumed to be calibrated. Indeed, the uncalibrated problem is ill-posed: the underlying normal map can be estimated only up to a linear ambiguity~\cite{Hayakawa1994}, which reduces to a generalized bas-relief one if integrability is enforced~\cite{Belhumeur1999}. To resolve the latter ambiguity, some prior on the scene's surface or geometry must be introduced, see~\cite{Shi2018} for a recent survey. A natural way to enforce integrability consists in following a differential approach to photometric stereo~\cite{Chand13,mecca14} i.e., directly estimate the 3D-surface as a depth map instead of first estimating the surface normals and then integrating them. Such a differential approach to photometric stereo can be coupled with variational methods in order to iteratively refine depth, reflectance and lighting in a robust manner~\cite{CVPR2017}. In addition to the theoretical interest of enforcing integrability in order to limit ambiguities, differential approaches to photometric stereo have the advantages of easing combination with other 3D-reconstruction methods~\cite{Gotardo2015,Peng2017}, and of bypassing the problem of integrating the estimated normal field, which is by itself a non-trivial problem~\cite{Queau_Survey}. Besides, any error in the estimated normal field might propagate during integration, and thus robustness to specularities or shadows must be enforced during normal estimation, see again~\cite{Shi2018} for some discussion.   
 
 All the research works mentioned in the previous paragraph assume that lighting is induced by a single light source. Nevertheless, many studies rather considered the case of more general illumination conditions, which finds a natural application in outdoor conditions~\cite{Sato1995}. For instance, the apparent motion of the sun within a day induces changes in the illumination direction which, in theory, allow photometric stereo-based 3D-reconstruction. However, this apparent motion is close to being planar, and thus the rank of the set of illumination vectors is equal or close to~$2$~\cite{Shen2014} (see also~\cite{Holdgeoffroy2015} for additional discussion on the stability of single-day photometric stereo). This situation is thus similar to the two-image case, which is known to be ill-posed since the early 90s~\cite{Kozera1993b,Onn1990,Yang1992}, although it is still an active research area~\cite{Kozera2018}. In order to limit the instabilities due to this issue, one possibility is to consider images acquired over many seasons as in~\cite{Abrams2012,Ackermann2012}, or to resort to deep neural networks~\cite{Holdgeoffroy2018}. Another one is to consider a non-directional illumination model to represent natural illumination, as for instance in~\cite{Jung2019}. Modeling natural illumination is a promising track, since such a model would not be restricted to sunny days, and images acquired under cloudy days are known to yield more accurate 3D-reconstructions~\cite{Holdgeoffroy2015}.

However, the previous approaches to photometric stereo under natural illumination assume calibrated lighting, where calibration is deduced from time and GPS coordinates or from a calibration target. The case of both general and uncalibrated lighting is much more challenging and has been fewly explored, apart from studies restricted to sparse 3D-reconstructions~\cite{Shen2009} or relying on the prior knowledge of a rough geometry~\cite{Ackermann2012b,Kemelmacher20103,Peng2017,Shi2014}. Uncalibrated photometric stereo under natural illumination has been revisited recently in~\cite{Mo2018}, using a spatially-varying equivalent directional lighting model. However, results were limited to the recovery of possibly non-integrable surface normals. Instead, the method which we propose in the present paper directly recovers the underlying surface represented as a depth map. Following the seminal work of Basri and Jacobs~\cite{Basri2007}, it considers the spherical harmonics representation of general lighting in lieu of the equivalent directional approximation, as discussed in the next section.

\section{Image Formation Model}
\label{sec:ifsha}

In photometric stereo (PS), we are given a number of observations $\{I^i\}_{i=1}^M$, each $I^i:\Omega\subset\bR^2\to \bR^C$ representing a multi-channel image (i.e.~$C\geq 1$) over a masked pixel domain $\Omega$. Assuming that the object being pictured is Lambertian, the surface's reflectance is represented by the albedo $\rho$, and the general image formation model is as follows, for all $i\in\{1,...,M\}$, $c\in\{1,...,C\}$, and $\fp\in\Omega$:
\ieqn{\label{eq:sfs}
\ialid{
I^i_c(\fp) = \int_{\bS^2} \rho_c(\fp)\ell^i_c(\bomega)\max\{\bomega\cdot\fn(\fp),0\}\diff{\bomega}. 
}}
Here $\bS^2$ is the unit sphere in $\bR^3$, $\ell^i_c:\bS^2\to\bR_+$ represents the channel-wise intensity of the incident light, and $\rho_c(\fp)\in\bR_+$ and $\fn(\fp)\in\bS^2$ are the channel-wise albedos and the unit-length surface normals, respectively, at the surface point conjugate to pixel $\fp\in\Omega$.
The $\max$ operation in~\eqref{eq:sfs} encodes self-shadows.
The overall integral $\int_{\bS^2}$ collects elementary luminance contributions arising from all incident lighting directions $\bomega$.
In the setup of uncalibrated PS, the quantities $\{\ell^i_c\}$, $\{\rho_c\}$, in addition to $\fn$, are unknown.

\emph{Equivalent directional lighting}~\cite{holdgeoffroy-15} approximates \eqref{eq:sfs} via
\ieqn{\label{eq:edl}
\ialid{
I^i_c(\fp) &= \rho_c(\fp)\, \bar\ell^i_c(\fp) \cdot \fn(\fp), \\
\bar\ell^i_c(\fp) &:= \int_{\{\bomega\in\bS^2:\,\bomega\cdot\fn(\fp)\geq0\}} \ell^i_c(\bomega)\bomega\diff{\bomega}.
}}
where $\bar\ell^i_c(\fp)$ represents the mean lighting over the visible hemisphere at $\fp$. The field $\bar\ell^c_i$ is \emph{spatially variant} but can be approximated by directional lighting over small local patches. Over each patch, one is thus faced with the ambiguities of directional uncalibrated PS~\cite{Hayakawa1994}. State-of-the-art patch-wise methods~\cite{Mo2018} first solve this problem over each patch, then connect the patches to form a complete normal field up to rotation, and eventually estimate the rotation which best satisfies the integrability constraint. Errors may however get propagated during the sequence, resulting in a possibly non-integrable normal field. 

\begin{figure}[t]
  \centering
  \newcommand{\mywidth}{0.1075\textwidth} 
  \newcolumntype{C}{ >{\centering\arraybackslash} m{0.02\textwidth} }
  \newcolumntype{X}{ >{\centering\arraybackslash} m{\mywidth} }
  \setlength\tabcolsep{1pt} 
  \def\arraystretch{1} 
  \begin{tabular}{XXXX}
  $\ell^i$ & Model \eqref{eq:sfs} & Model \eqref{eq:shem} (1st order) & Model \eqref{eq:shem} (2nd order) \\
    \includegraphics[width=\mywidth]{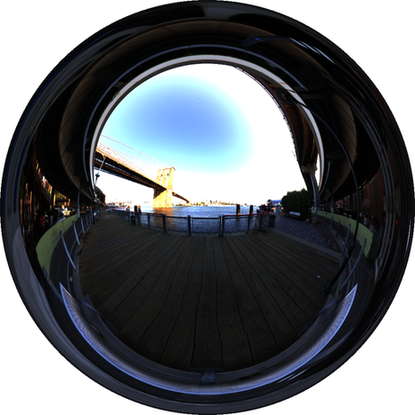}&
    \includegraphics[width=\mywidth]{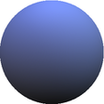}&
    \includegraphics[width=\mywidth]{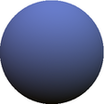}&
    \includegraphics[width=\mywidth]{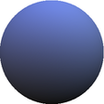}\\
    \includegraphics[width=\mywidth]{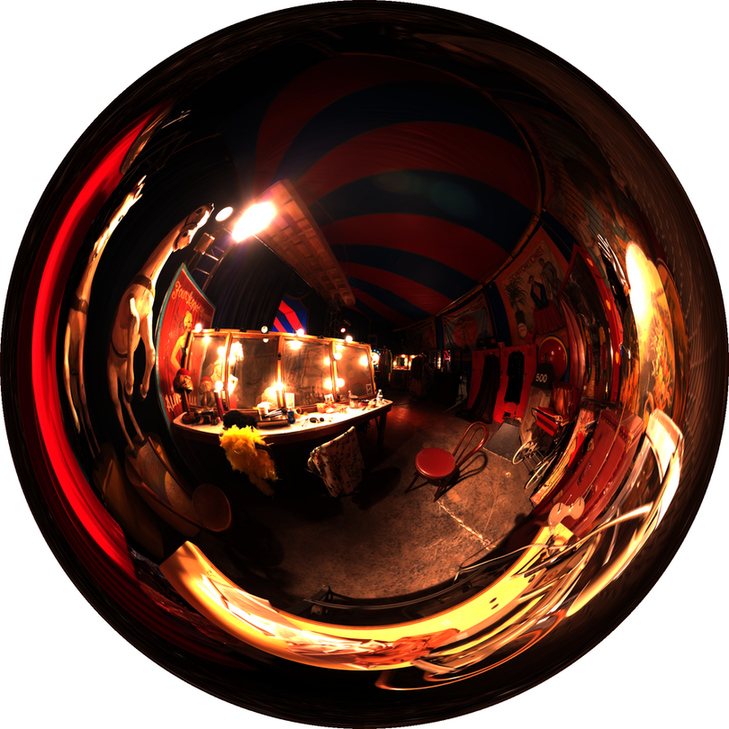}&
    \includegraphics[width=\mywidth]{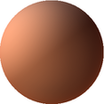}&
    \includegraphics[width=\mywidth]{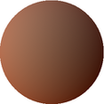}&
    \includegraphics[width=\mywidth]{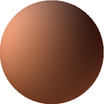}
  \end{tabular}
  \caption{Illustration of RGB ($C=3$) environment lighting $\ell^i=(\ell^i_1, \ell^i_2,\ell^i_3)$, the resulting images (assuming white albedos and a spherical shape) under the image formation model~\eqref{eq:sfs} and its approximation by spherical harmonics. 
  The approximation by the second-order spherical harmonics is nearly perfect.
}
\label{fig:sha}
\end{figure}

Instead of such an equivalent directional lighting model, we rather consider a \emph{spherical harmonic approximation} (SHA) of general lighting~\cite{Basri2003,Basri2007}. By defining the {half-cosine kernel} $k$ as 
\iali{
k(\bomega,\fn) &:= \max\{\bomega\cdot \fn,0\},
}
we can view \eqref{eq:sfs} as an analog of a convolution:
\iali{
I^i_c(\fp) &= \rho_c(\fp) \int_{\bS^2} k(\bomega,\fn(\fp)) \ell^i_c(\bomega)\diff{\bomega}.
\label{eq:sfsc}
}
Invoking the Funk-Hecke theorem, we obtain the following harmonic expansion analogous to Fourier series:
\iali{
& \int_{\bS^2} k(\bomega,\fn(\fp)) \ell^i_c(\bomega)\diff{\bomega} = \sum_{n=0}^\infty\sum_{m=-n}^n (k_n\ell_{n,m}^{i,c}) h_{n,m}(\fn(\fp)). 
\label{eq:she}
}
Here the spherical harmonics $\{h_{n,m}\}$ form an orthonormal basis of $L^2(\bS^2)$, and $\{k_n\}$ and $\{\ell_{n,m}^{i,c}\}$ are the expansion coefficients of $k$ and $\ell^i_c$ with respect to $\{h_{n,m}\}$. Since most energy in the expansion \eqref{eq:she} concentrates on low-order terms~\cite{Basri2003}, we obtain the \emph{second-order SHA} by truncating the series up to the first nine terms (i.e., $0\leq n\leq2$):
\iali{
& \int_{\bS^2} k(\bomega,\fn(\fp)) \ell^i_c(\bomega)\diff{\bomega} \approx \sum_{n=0}^2\sum_{m=-n}^n (k_n\ell_{n,m}^{i,c}) h_{n,m}(\fn(\fp)).
\label{eq:she2}
}
The first-order SHA refers to the truncation up to the first four terms (i.e., $0\leq n\leq1$).  It is shown in \cite{Basri2003} that, for distant lighting, at least $75\%$ of the resulting irradiance is captured by the first-order SHA, and $98\%$ by the second-order SHA (cf. Figure~\ref{fig:sha} for a visualization).

Plugging \eqref{eq:she2} and specifics of spherical harmonics \cite{Basri2003} into \eqref{eq:sfsc}, we finalize our image formation model as:
\iali{
&\!\!\!I^i_c(\fp) \approx \rho_c(\fp)\, 
\fl^i_c\cdot\fh[\fn](\fp), \label{eq:shem}\\
&\!\!\!\fh[\fn] \!=\!
\imtx{
\!1,\!
\fn_1,\fn_2,\!
\fn_3,\!
\fn_1\fn_2,\!
\fn_1\fn_3,\!
\fn_2\fn_3, \!
\fn_1^2\!-\!\fn_2^2,\!
3\fn_3^2\!-\!1\!
}^{\!\top}\!\!\!.
}
Here $\fh[\fn]:\Omega\to\bR^9$ represents the second-order harmonic images, and $\fl^i_c\in\bR^9$ represents the harmonic lighting vector whose entries have absorbed $\{k_n\ell_{n,m}^{i,c}\}$ and constant factors of $\{h_{n,m}\}$. 
A key advantage of the SHA~\eqref{eq:shem} over the equivalent directional lighting model \eqref{eq:edl} lies in the \emph{spatial invariance} of the lighting vectors $\{\fl^i_c\}$, which yields a less ill-posed inverse problem~\cite{Basri2007}. The counterpart is the nonlinear dependency upon the normal components, which we will handle in Section~\ref{sec:5} using a tailored numerical solution. In the next section, we build upon the key observations that integrability~\cite{Belhumeur1999} and perspective projection~\cite{papadhimitri2013new} both largely reduce the ambiguities of uncalibrated PS to derive a variational approach to inverting the SHA~\eqref{eq:shem}.

\section{Variational Uncalibrated PS}
\label{sec:vups}

In this section, we shall propose a joint variational model for uncalibrated PS. 
To this end, let a 3D-frame $(Oxyz)$ be attached to the camera, with $O$ the optical center, the $z$-axis aligned with the optical axis such that $z>0$ for any 3D point $(x,y,z)$ in front of the camera.
Further let a 2D-frame $(O'uv)$ be attached to the focal plane which is parallel to the $xy$-plane and contains the masked pixel domain $\Omega$. Under perspective projection, the surface geometry is modeled as a map $\fx:\fp=(u,v)\in\Omega\mapsto \fx(u,v)\in\bR^3$ given by
\iali{
\fx(u,v) &= z(u,v)K^{-1}[u,v,1]^\top,
}
with $z:\Omega\to\bR_+$ the \emph{depth} map  and 
\iali{
K:=\imtx{
f_u & 0 & u_0 \\
0 & f_v & v_0 \\
0 & 0 & 1
}
}
the calibrated camera's intrinsics matrix. In the following we denote for convenience $(\tilde{u},\tilde{v}):=(u-u_0,v-v_0)$.

Assuming that $z$ is differentiable, the surface normal $\fn$ at point $\fx(u,v)$ is the unit vector oriented towards the camera such that
$\fn(u,v) \propto \partial_u\fx(u,v) \times \partial_v\fx(u,v)$, 
which yields the following parameterization of the normal by the depth:
\iali{
\fn[z](u,v) &= \frac{\tn[z](u,v)}{|\tn[z](u,v)|}, \label{eq:nvd}\\
\tn[z](u,v) &:= \imtx{
f_u \partial_u z(u,v) \\
f_v \partial_v z(u,v) \\
-z(u,v) -\tilde{u}\, \partial_u z(u,v) - \tilde{v}\,\partial_v z(u,v)
}.
}
Note that the dependence of $\tn[z]$ on $z$ is linear.

Based on the forward model \eqref{eq:shem} and the parameterization \eqref{eq:nvd} of normals, we formulate the joint recovery of reflectance, lighting and geometry as the following variational problem: 
\iali{
&\hspace{-.3cm}\min_{\{\rho_c\}, \{\fl^i_c\}, z} \sum_{i=1}^M\sum_{c=1}^C\int_\Omega \phi_\lambda\Big(\rho_c(u,v)\,\fl^i_c\cdot \fh[\fn[z]](u,v) \notag\\[-0.5em]
& 
-I^i_c(u,v)\Big) \diff{u}\diff{v} +\mu\sum_{c=1}^C\int_\Omega |\nabla\rho_c(u,v)|_\gamma \diff{u}\diff{v}. \label{eq:cvm}
}
In the first term above, we use \emph{Cauchy's M-estimator} to penalize the data-fitting discrepancy:
\iali{
\phi_\lambda(s) &= \lambda^2\log(1+s^2/\lambda^2),
}
It is indeed well-known that Cauchy's estimator, being non-convex, is robust against outliers; see for instance~\cite{CVPR2017} in the context of PS. The scaling parameter $\lambda=0.15$ is used in all experiments.

The second term in \eqref{eq:cvm} represents a Huber total-variation (TV) regularization on each albedo map $\rho_c$, with the Huber loss defined by
\iali{
|s|_\gamma &:= 
\begin{cases}
|s|^2/(2\gamma) & \text{if }|s|\leq\gamma,\\
|s|-\gamma/2 & \text{if }|s|>\gamma,
\end{cases}
}
and $\gamma=0.1$ being fixed in the experiments. It turns out that the Huber TV imposes desirable smoothness on the albedo maps $\{\rho_c\}$ and in turn improves the joint estimation overall. Eventually, $\mu>0$ is a weight parameter which balances the data-fitting term and the Huber TV one. Its value was empirically set to $2\cdot 10^{-6}$ (see Section~\ref{sec:6} for some discussion). 

In \eqref{eq:cvm}, geometry is directly optimized in terms of the depth $z$ (rather than indirectly in terms of the normal $\fn$). This both ensures integrability and avoids integration of normals into depths as a post-processing step.
\section{Solver and Implementation}\label{sec:5}
To solve the variational problem \eqref{eq:cvm} numerically, we follow a ``discretize-then-optimize'' approach. There, $\Omega \subset \bR^2$ is replaced by $\bR^N$, $N$ being the number of pixels inside $\Omega$,
which yields discretized vectors $z,\{\rho_c\}_{c=1}^C \in \bR^N$.
To alleviate notational burden, we sometimes refer to a pixel by its index $j \in \{1,\dots,N\}$ and sometimes by its position $\fp=(u,v)\in\Omega$.
The spatial gradient $\nabla$ is discretized using a forward difference stencil.

We shall apply a lagged block coordinate descent (LBCD) method to find a local minimum of the objective function in~\eqref{eq:dcvm}. Due to the (highly) non-convex nature of~\eqref{eq:dcvm}, initialization of optimization variables has a strong influence on the final solution. In our implementation, we initialize $\rho_{c,j} = \text{median}(\{I^i_{c,j}\}_{i=1}^M)$ for all $c,j$ and $\fl_c^i = [0.2, 0, 0, -1, 0, 0, 0, 0, 0]^\top$ for all $c,i$. Moreover, during the first eight iterations we freeze the second-order spherical harmonics coefficients $(\fl_c^i)_5=(\fl_c^i)_6=...=(\fl_c^i)_9=0$ i.e., we reconstruct using only first-order spherical harmonic approximation as a warm start. Most real-world scenes being convex, we initialize the depth $z$ as a balloon-like surface, as discussed in the following.

\subsection{Depth Initialization}\label{sec:5.1}

\begin{figure}[h!]
  \centering
  \newcommand{\mywidth}{0.115\textwidth}
  \newcommand{\mywidthy}{0.09\textwidth}
  \newcolumntype{X}{ >{\centering\arraybackslash} m{\mywidth} }
  \setlength\tabcolsep{1pt} 
  \def\arraystretch{1} 
  \begin{tabular}{XXXX}
    \includegraphics[width=\mywidthy]{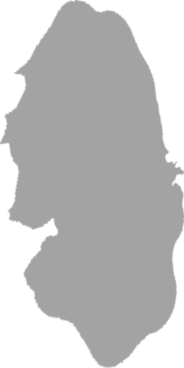}&
    \includegraphics[width=\mywidth]{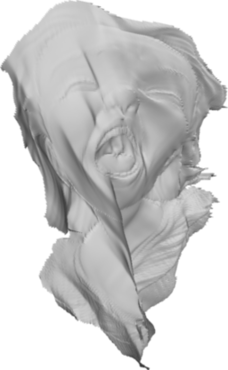}&
    \includegraphics[width=\mywidthy]{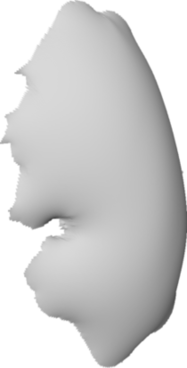}&
    \includegraphics[width=\mywidth]{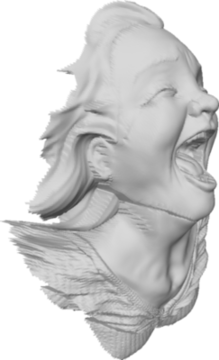}\\
      \multicolumn{2}{c}{Trivial $z_p\equiv1$ and its result $z$} & \multicolumn{2}{c}{Our $z_p$ and its result $z$} \vspace{0.5cm}    
  \end{tabular}\caption{Impact of depth initialization: a trivial constant initialization on the left vs. our initialization on the right and its corresponding resulting geometry estimates. Further results from varying initializations can be found in the supplementary material.}
\label{fig:balloon_init}
\end{figure}

It is readily seen that a trivial constant initialization of the depth $z$ yields uniform vertically aligned normals $\fn[z]$ and, hence, zero entries in the initial harmonic images $\fh[\fn[z]]$. This would cause non-meaningful updates on albedos $\{\rho_c\}$ and lighting vectors $\{\fl_c^i\}$;
cf.~Figure \ref{fig:balloon_init} for an illustration.

To solve this issue, we specialize the depth initialization which undergoes two phases:

1. Following \cite{Oswald2012}, we generate a balloon-like depth map $z_o$ under orthographic projection.

2. We then convert the orthographic depth $z_o$ to a perspective depth $z_p$ via normal integration \cite{Queau_Survey}.

Phase 1 is pursued via seeking a depth map $z_o$ which has minimal surface area subject to a constant volume $V$:
\begin{equation}
	\begin{aligned}
		\min_{z_o} &\int_\Omega \sqrt{1+|\nabla z_o|^2} \diff{u}\diff{v}\\
		\st &\int_{\Omega} z_o \diff{u}\diff{v} = V.
	\end{aligned}\label{eq:balloon}
\end{equation}
A global minimizer of this model can be efficiently computed by simple projected gradient iterations:
\begin{align}
	z_o^{(k+1/2)} &= z_o^{(k)} - \tau \nabla^\top \left(\frac{1}{\sqrt{1+|\nabla z_o^{(k)}|^2}}\nabla z_o^{(k)}\right) \label{eq:balloon_gd},\\
	z_o^{(k+1)} &= z_o^{(k+1/2)} + \left(\frac{V - \int_{\Omega} z_o^{(k+1/2)} \diff{u}\diff{v}}
	{\int_{\Omega}\diff{u}\diff{v}} \right) \cdot \textbf{1}_{\Omega} \label{eq:ballon_pj},
\end{align}
where $\textbf{1}_\Omega(u,v)\equiv 1$ and $\tau=0.8/\|\nabla\|_\text{spec}$ with $\|\cdot\|_\text{spec}$ the spectral norm. The volume constant $V$ is a hyperparameter which is empirically chosen, see Section~\ref{sec:6} for discussion.

Next, we convert the orthographic depth $z_o$ to a perspective depth $z_p$.
Note that $z_o$ complies with the orthographic projection, under which a 3D-point $\hat\fx$ is represented by
\begin{equation}
	\hat\fx(u,v) = [u, v, z_o(u,v)]^\top,
\end{equation}
and the corresponding surface normal $\hat\fn$ to the surface at $\hat\fx$ conjugate to pixel $\hat\fp=(u,v)$ is given by
\begin{equation}
	\hat\fn(u,v) = \frac{1}{\sqrt{|\nabla z_o(u,v)|^2+1}}[ \nabla z_o(u,v), -1]^\top.
	\label{eq:onvd}
\end{equation}
Since $\hat\fn$ is invariant to the projection model, Eq.~\eqref{eq:nvd} also implies that
\begin{equation}
	\hat\fn(u,v) \propto \imtx{
		f_u \partial_u \tz_p(u,v) \\
		f_v \partial_v \tz_p(u,v) \\
		-1 -\tilde{u}\partial_u \tz_p(u,v) - \tilde{v}\partial_v \tz_p(u,v)},
\end{equation}
where $\tz_p(u,v)=\log z_p(u,v)$ stands for the log-perspective depth. This further implies the formula for $\nabla\tz_p$:
\iali{
\nabla\tz_p(u,v) &= \frac{-1}{\frac{\tilde{u}\hat\fn_1(u,v)}{f_u}+\frac{\tilde{v}\hat\fn_2(u,v)}{f_v}+\hat\fn_3(u,v)}\imtx{\frac{1}{f_u}\hat\fn_1(u,v) \\ \frac{1}{f_v}\hat\fn_2(u,v)},
\label{eq:lpnvd}
}
which can be integrated to obtain $\tz_p$ (and hence $z_p$). The overall pipeline in Phase 2 is summarized as follows:
\begin{enumerate}
	\item ($z_o \to \hat\fn$): Compute $\hat\fn$ by \eqref{eq:onvd}.
	\item ($\hat\fn \to \nabla\tz_p$): Compute $\nabla\tz_p$ by \eqref{eq:lpnvd}.
	\item ($\nabla\tz_p \to z_p$): Perform integration~\cite{Queau_Survey} to obtain $\tz_p$. Return $z_p=\exp\tz_p$ as the initialized (perspective) depth.
\end{enumerate}
As discussed in~\cite{Graber2015} the perspective surface area depends linearly on the depth $z$. This complicates direct perspective ballooning, since the depth is driven towards zero and hence yields numerical instability. For this reason, we opted for the two-step approach which bypasses the issue.

\subsection{Lagged Block Coordinate Descent}
Even with a reasonable initialization, the numerical resolution of Problem \eqref{eq:dcvm} remains challenging. Due to the appearances of the spherical harmonic approximation $\fh[\fn[z]]$ and the Cauchy's M-estimator $\phi_\lambda$, the objective in \eqref{eq:dcvm} is highly nonlinear and nonconvex. To tackle these challenges, here we present a lagged block coordinate descent (LBCD) method which performs efficiently in practice.

To derive LBCD, we introduce an auxiliary variable $\theta\in\bR^N$ such that $\theta_j=| \tn_j[z]|$. This enables us to rewrite~\eqref{eq:nvd} as $\fn_j[z] = \tn_j[z]/\theta_j$.
Then we formulate the following constrained optimization problem:
\begin{equation}
	\begin{aligned}
		\min_{\theta,\{\rho_c\}, \{\fl_c^i\}, z}&
	\sum_{i=1}^M \sum_{c=1}^C \sum_{j=1}^N \phi_\lambda \left(r_{i,c,j}(\theta_j, \rho_{c,j}, \fl_c^i, z)\right) \\
	&+ \mu \sum_{c=1}^C \sum_{j=1}^N|(\nabla \rho_c)_j|_\gamma, \\
	\st ~& \theta_j = |\tn_j[z]|, ~\forall j \in \{1,\dots,N \},
	\end{aligned} 
	\label{eq:dcvm}
\end{equation}
where $r_{i,c,j}$ is the residual function defined by:
\begin{equation}
	r_{i,c,j}(\theta_j, \rho_{c,j}, \fl_c^i, z) =  \rho_{c,j}\, \fl_c^i \cdot \fh_j [\tn_j[z] / \theta_j] - I_{c,j}^i.
\end{equation}

Upon initialization, the proposed LBCD proceeds as follows. At iteration $k$, we lag $\theta$ one iteration behind, i.e.,
\iali{
\theta_j^{(k+1)} &:= |\tn_j[z^{(k)}]|, \quad\forall j\in\{1,...,N\},
}
and then sequentially update each of the three blocks (namely $\{\rho_c\}$, $\{\fl_c^i\}$ and $z$). In each resulting subproblem, we solve (lagged) weighted least squares problems as an approximation of the Cauchy loss and/or the Huber loss. This is detailed in the following:

\begin{itemize}
	\item (Update $\{\rho_c\}$):
		We evaluate the residual
		\iali{
		r_{i,c,j}^{(k+1/3)} := r_{i,c,j}(\theta_j^{(k+1)}, \rho_{c,j}^{(k)}, \fl_{c}^{i,(k)}, z^{(k)}),
		}
		and then set up the (lagged) weight factors for both the Cauchy loss and the Huber loss as
		\iali{
			w_{i,c,j}^{(k+1/3)} &:= 
			{\phi^\prime_\lambda(r_{i,c,j}^{(k+1/3)})}/{r_{i,c,j}^{(k+1/3)}}, \\
			\label{eq:reweight}
			q_{c,j}^{(k+1/3)} &:= 1/\max\{\gamma, |(\nabla \rho^{(k)}_c)_j|\}.
		}
		The albedos $\{\rho_c\}$ are updated as the solution to the following linear weighted least-squares problem:
		\begin{equation}
			\begin{aligned}
				&\{\rho_c^{(k+1)}\} := \arg\min_{\{\rho_c\}}  \mu \sum_{c,j}q_{c,j}^{(k+1/3)} |(\nabla \rho_c)_j|^2 \\
				&\quad+\sum_{i,c,j} w_{i,c,j}^{(k+1/3)}|r_{i,c,j}(\theta_j^{(k+1)}, \rho_j^c, \fl_{c}^{i,(k)}, z^{(k)})|^2,
		\end{aligned}
		\end{equation}
		which is carried out by conjugate gradient (CG).
			
	\item (Update $\{\fl_c^i\}$):
		The lighting subproblem is similar to the one for albedos, except for absence of the Huber TV term. Upon evaluation of the residual $r_{i,c,j}^{(k+2/3)}$ and the weight factor $w_{i,c,j}^{(k+2/3)}$, we update $\{\fl_c^i\}$ by solving the following linear weighted least-squares problem via CG:
		\ieqn{
		\ialid{
			&\{\fl_{c}^{i,(k+1)}\} = \arg\min_{\fl_c^i} \sum_{i,c,j} w_{i,c,j}^{(k+2/3)}\cdot \\
			&\qquad|r_{i,c,j}(\theta_j^{(k+1)}, \rho_{c,j}^{(k+1)}, \fl_c^i, z^{(k)})|^2. 
		}}
			
	\item (Update $z$): 
		The depth subproblem requires additional efforts. With $r_{i,c,j}^{(k+1)}$ and $w_{i,c,j}^{(k+1)}$ evaluated after the $\{\fl_c^i\}$-update, we are faced with the following weighted least squares problem:
		\ieqn{
		\ialid{
			& \min_z \sum_{i,c,j} w_{i,c,j}^{(k+1)} |r_{i,c,j}(\theta^{(k+1)}_j, \rho^{(k+1)}_{c,j}, \fl_{c}^{i,(k+1)}, z)|^2,
		}}
		where the dependence of $r_{i,c,j}$ on $z$ is still nonlinear. Therefore, we further linearize $r_{i,c,j}$ with respect to $z$ and arrive at the following update:
		\ieqn{\ialid{
		&z^{(k+1)} = \arg\min_z \sum_{i,c,j} w_{i,c,j}^{(k+1)}\cdot  \\
		&\qquad |r_{i,c,j}^{(k+1)} + J_r(z^{(k)})(z-z^{(k)})|^2, 
		}}
		where $J_r(z^{(k)})$ is the Jacobian of the map $z\mapsto r_{i,c,j}(\theta_j^{(k+1)}, \rho_{c,j}^{(k+1)}, \fl_{c}^{i,(k+1)}, z)$ at $z=z^{(k)}$. The resulting linearized least-squares problem is again solved by CG. In our experiments, we additionally incorporate backtracking line search in the $z$-update to ensure a monotonic decrease of the energy.
\end{itemize}
\section{Experimental Validation}\label{sec:6}

\begin{figure*}[ht!]
  \centering
  \newcommand{\mywidth}{0.09\textwidth} 
  \newcommand{\mywidthy}{0.175\textwidth} 
  \newcolumntype{C}{ >{\centering\arraybackslash} m{0.02\textwidth} }
  \newcolumntype{X}{ >{\centering\arraybackslash} m{\mywidth} }
  \newcolumntype{Y}{ >{\centering\arraybackslash} m{\mywidthy} }
  \setlength\tabcolsep{4pt} 
  \def\arraystretch{1} 
  \begin{tabular}{CXXXXXXXXX}
    \rotatebox{90}{3D Shapes}&
    \includegraphics[width=\mywidth]{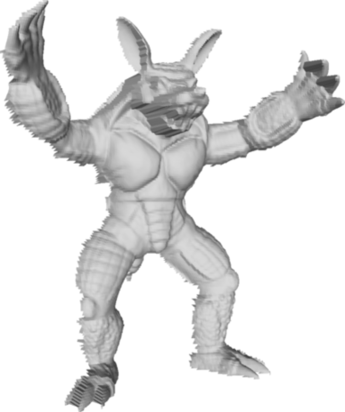}&
    \includegraphics[width=\mywidth]{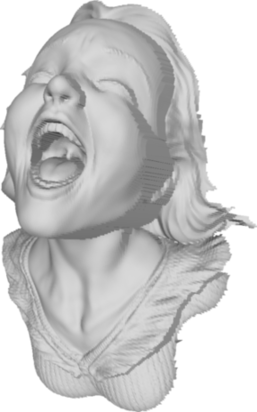}&
    \includegraphics[width=\mywidth]{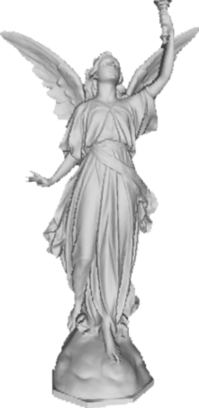}&
    \includegraphics[width=\mywidth]{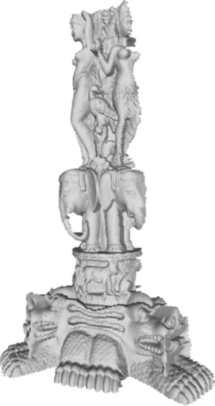}&
    \multicolumn{4}{|Y}{\includegraphics[width=\mywidthy]{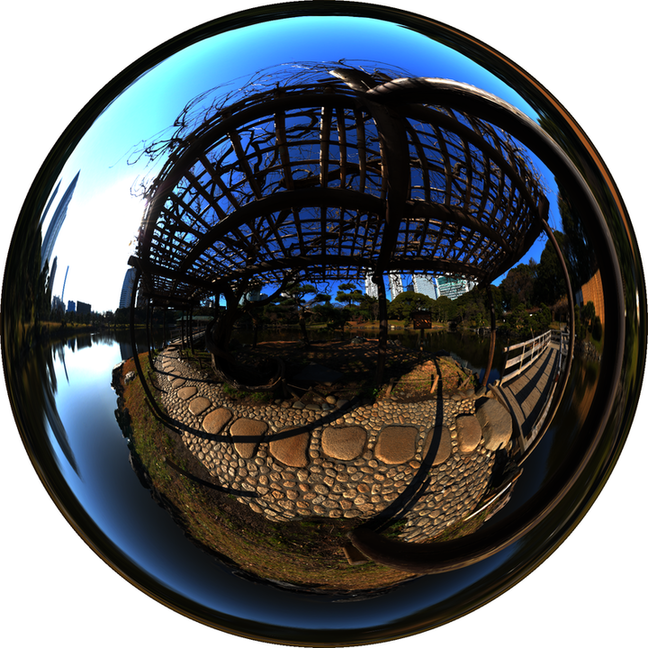}%
    \includegraphics[width=\mywidthy]{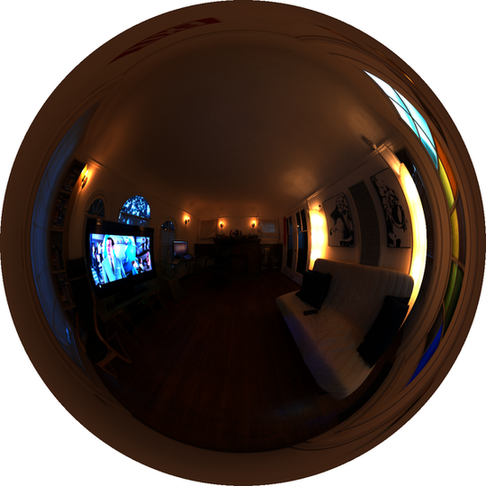}%
    \includegraphics[width=\mywidthy]{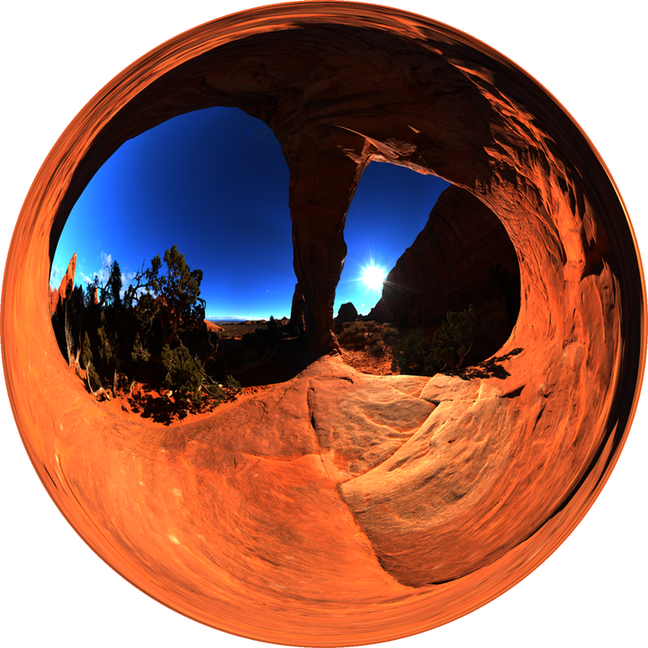}}
    \\
     &Armadillo&
    Joyful Yell&
    Lucy&
    Thai Statue&
    \multicolumn{5}{|c}{Exemplary environment maps}\\
    \hline
    \vspace{1.1cm}
    \rotatebox{90}{Albedos}&
    \includegraphics[width=\mywidth]{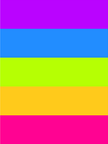}&
    \includegraphics[width=\mywidth]{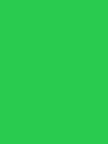}&
    \includegraphics[width=\mywidth]{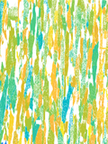}&
    \includegraphics[width=\mywidth]{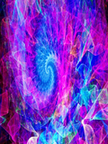}&
    \includegraphics[width=\mywidth]{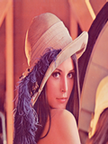}&
    \includegraphics[width=\mywidth]{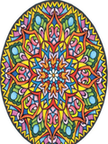}&
    \includegraphics[width=\mywidth]{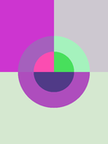}&
    \includegraphics[width=\mywidth]{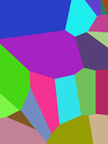}&
    \includegraphics[width=\mywidth]{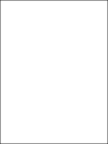}\\
        &Bars&
    Constant&
    Ebsd&
    Hippie&
    Lena&
    Pattern&
    Rectcircle&
    Voronoi&
    White
  \end{tabular}
  \caption{The four 3D-shapes and nine albedo maps we used to create 36 (3D-shape, albedo) datasets. For each dataset, $M=25$ images were rendered using different environment maps such as those shown on the top right. }
  \label{fig:illustration_synthetic_data}
\end{figure*}

This section is concerned with the evaluation of the proposed nonconvex variational approach to uncalibrated photometric stereo under general lighting.

\subsection{Synthetic Experiments}
To validate the impact of the initial volume $V$ in \eqref{eq:balloon}, the tunable hyper-parameter $\mu$, and the number of input images~$M$ in \eqref{eq:cvm}, we consider 36 challenging synthetic datasets. We use four different depth maps (``Joyful Yell''~\cite{Bendansie2015}, ``Lucy''~\cite{Levoy2005data}, ``Armadillo''~\cite{Levoy2005data} and ``Thai Statue''~\cite{Levoy2005data}) and nine different albedo maps and each of those 36 combinations is rendered as described in \eqref{eq:sfs} using $M=25$ different environment maps\footnote{Environment maps are downloaded from \url{http://www.hdrlabs.com/sibl/archive.html}}, cf. Figure~\ref{fig:illustration_synthetic_data}. The resulting 25 RGB images per dataset are used as input, along with the intrinsic camera parameters and a binary mask $\Omega$. A quantitative evaluation on the triplet $(V, \mu, M)$ is carried out on four randomly chosen datasets (Armadillo \& White albedo, Joyful Yell \& Ebsd albedo, Lucy \& Hippie albedo, and Thai Statue \& Voronoi albedo), comparing the impact of each value of $(V, \mu, M)$ on the resulting mean angular error (MAE) between ground truth and estimated normals.

\begin{figure}[!ht]
  \centering
  \newcommand{\mywidth}{0.23\textwidth}
  \newcolumntype{X}{ >{\centering\arraybackslash} m{\mywidth} }
  \setlength\tabcolsep{1pt} 
  \def\arraystretch{1} 
  \begin{tabular}{XX}
    \includegraphics[width=\mywidth]{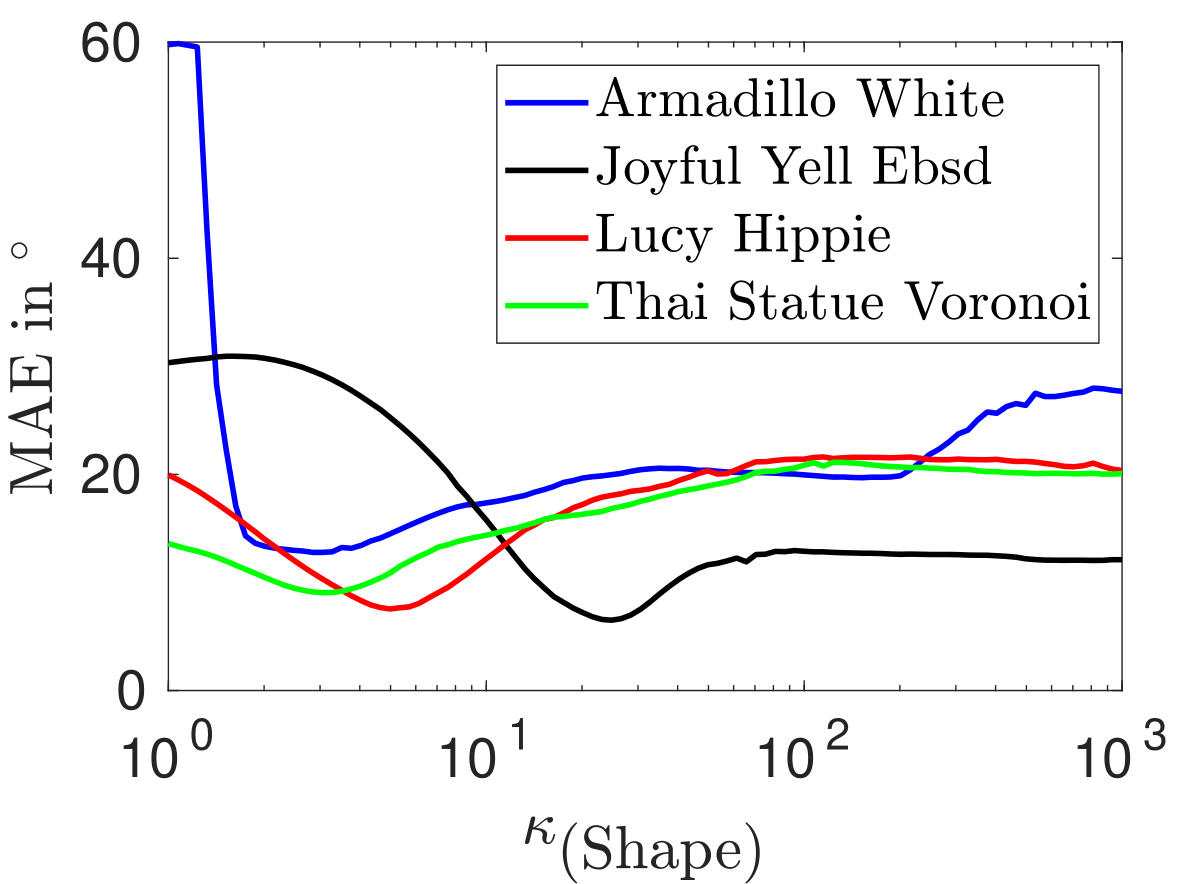}&
    \includegraphics[width=\mywidth]{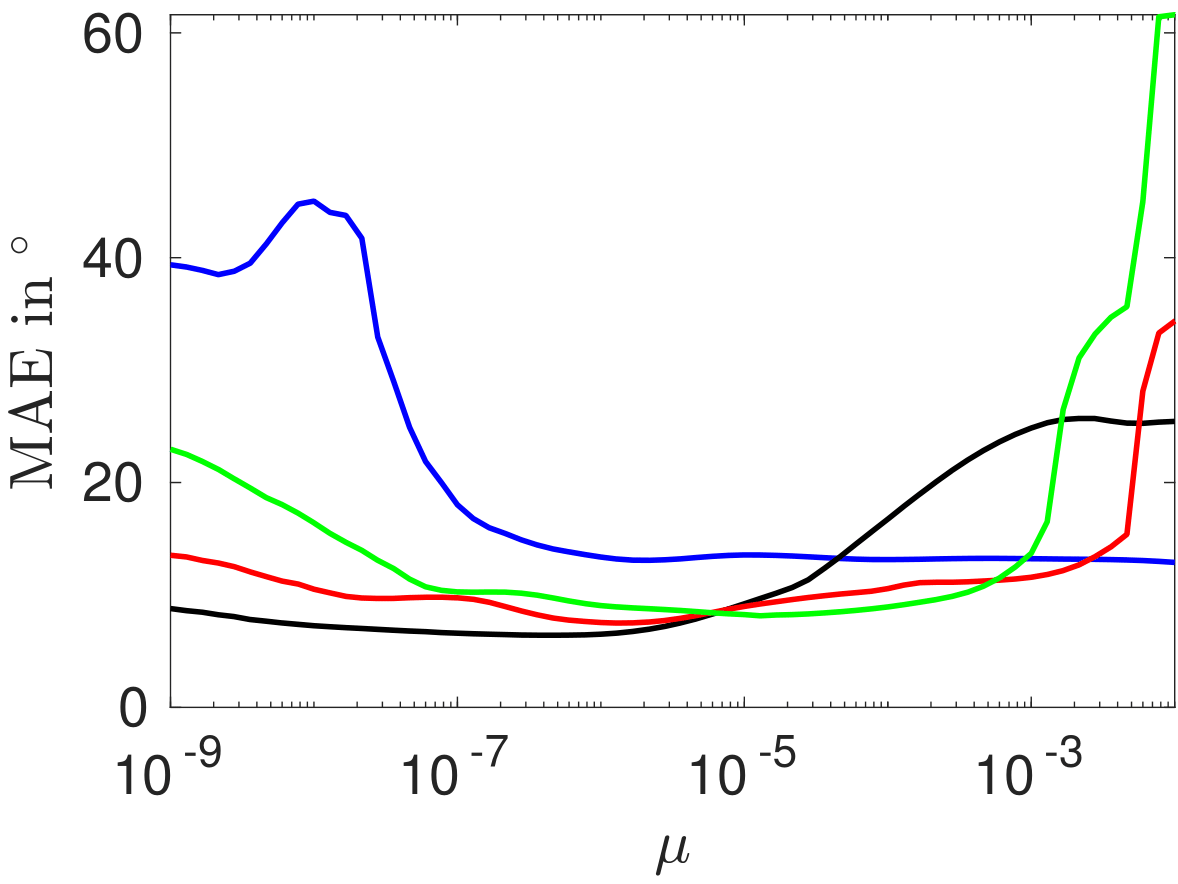}
  \end{tabular}
  \caption{Impact of the initial volume $V_\text{(Shape)}$ as well as $\mu$ on the accuracy of the estimated depth. Based on these experiments we choose $\kappa_\text{(Armadillo)} = 2.84$, $\kappa_\text{(Joyful Yell)} = 24.77$, $\kappa_\text{(Lucy)} = 4.98$, $\kappa_\text{(Thai Statue)} = 3.05$ and $\mu=2\cdot10^{-6}$ for all experiments, where $V_\text{(Shape)} = \kappa_\text{(Shape)}N_\text{(Shape)}$.
  }
  \label{fig:volume_tuning}
\end{figure}

First, we validate the choice of the input volume $V$ using the initially fixed values of $\mu=2\cdot10^{-6}$ and $M=25$. As the volume depends on the size of the mask, we consider a linear parametrization $V(\kappa)=\kappa |\Omega|=\kappa N$ and evaluate a range of ratios $\kappa\in\left[1, 10^3\right]$. Figure~\ref{fig:volume_tuning} (left) indicates that the optimal value of $\kappa$ is dataset-dependent. For synthetic datasets we always selected this optimal value, yet for real-world data no such evaluation is possible and $\kappa$ must be tuned manually. Since the ballooning-based depth initialization can be carried out in real-time (implementation is parallelized in CUDA), the user has an immediate feedback on the initial depth and thus a plausible initial shape is easily drawn. Humans excel at estimating size and shape of objects~\cite{Baldwin2016} and real-world experiments will show that a manual choice of $\kappa$ can result in appealing geometries.

Next, we evaluate the impact of $\mu$, cf. Figure~\ref{fig:volume_tuning} (right). As can be seen, the depth estimate seems to deteriorate for too small and too large values of $\mu$, whereas $\mu\in\left[10^{-6},10^{-5}\right]$ seems to provide good depth estimates across all albedo maps. Therefore we fix $\mu=2\cdot10^{-6}$ for all our upcoming experimental evaluation. 

\begin{figure}[!ht]
  \centering
  \newcommand{\mywidth}{0.23\textwidth}
  \newcolumntype{X}{ >{\centering\arraybackslash} m{\mywidth} }
  \setlength\tabcolsep{1pt} 
  \def\arraystretch{1} 
  \begin{tabular}{XX}
    \includegraphics[width=\mywidth]{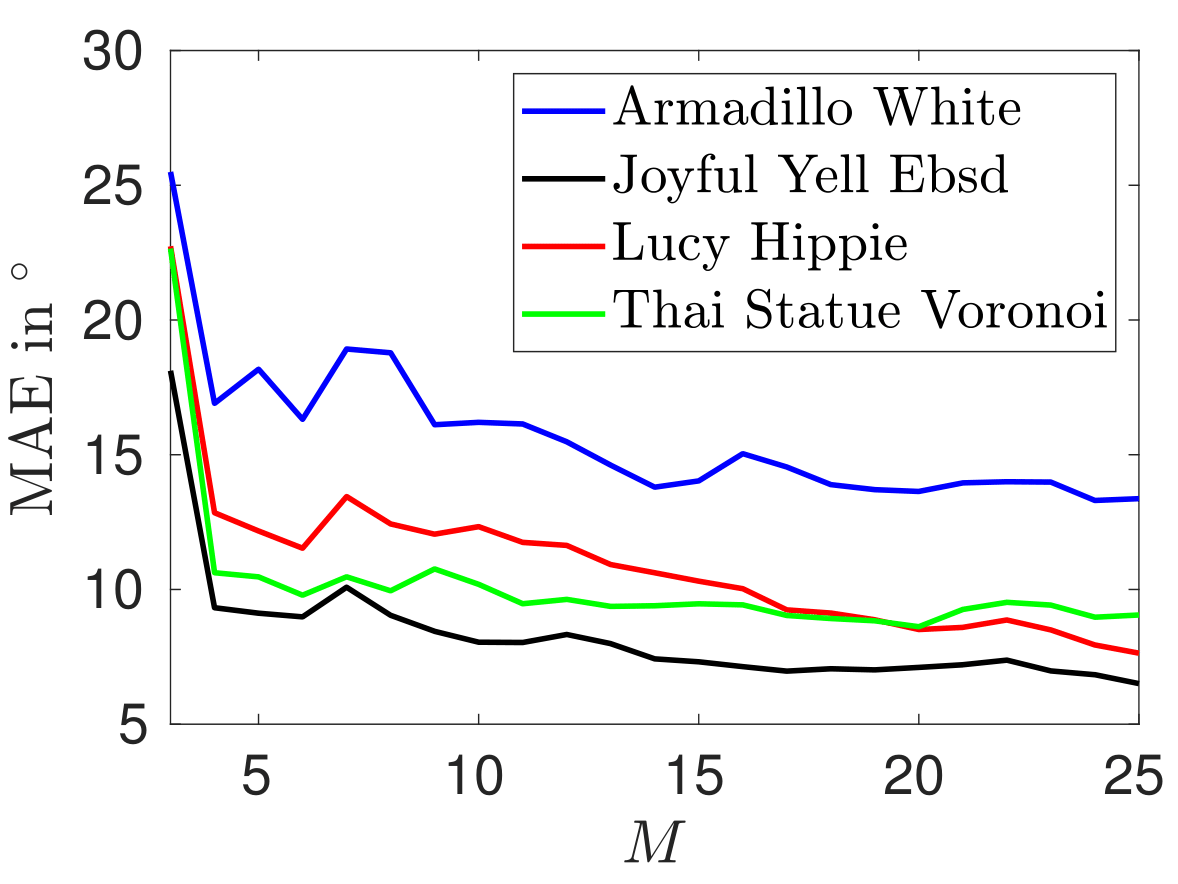}&
    \includegraphics[width=\mywidth]{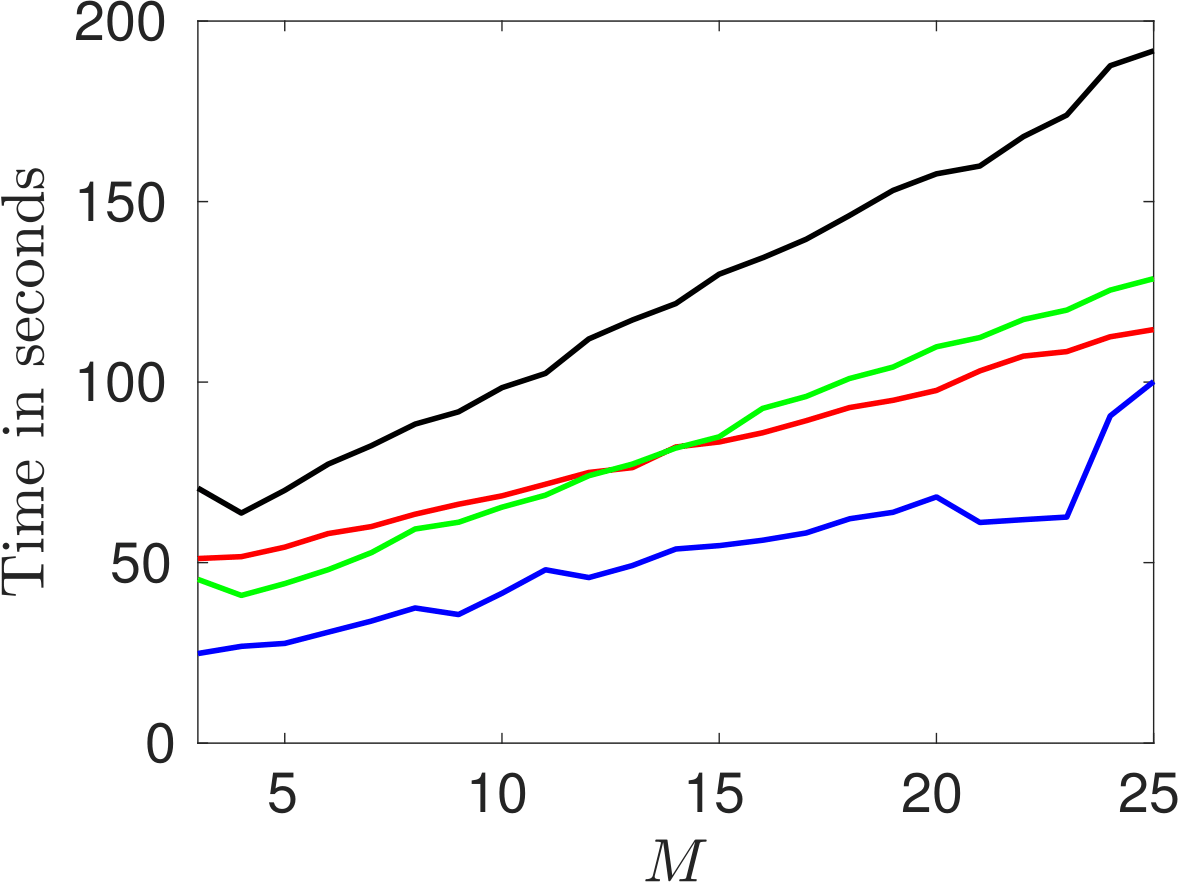}
  \end{tabular}
  \caption{Impact of the number of images $M$ on the mean angular error (MAE) and the runtime. Based on these insights we choose $M=20$ for our experiments.}
  \label{fig:param_tuning}
\end{figure}

Unsurprisingly, the MAE is inversely proportional to the number $M$ of input images, but runtime increases (linearly) with $M$, cf. Figure~\ref{fig:param_tuning}. We found that $M\in\left[15,25\right]$ represents a good trade-off between runtime and accuracy, and fix $M=20$ for all our further experiments. Our Matlab implementation needs about $1$--$2$ minutes on a computer with an Intel $i7$ processor.

\begin{figure}[!ht]
  \centering
  \newcommand{\mywidth}{0.07175\textwidth} 
  \newcolumntype{C}{ >{\centering\arraybackslash} m{0.02\textwidth} }
  \newcolumntype{X}{ >{\centering\arraybackslash} m{\mywidth} }
  \setlength\tabcolsep{1pt} 
  \def\arraystretch{1} 
  \begin{tabular}{CXXXXXX}
    &$I^i$&\cite{Favaro2012}&\cite{Peng2017}&\cite{Mo2018}& {Ours}& GT\\\hline
    \rotatebox{90}{Joyful Yell \& Lena}&
    \includegraphics[width=\mywidth, height=50pt]{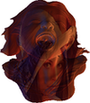}&
    \includegraphics[width=\mywidth]{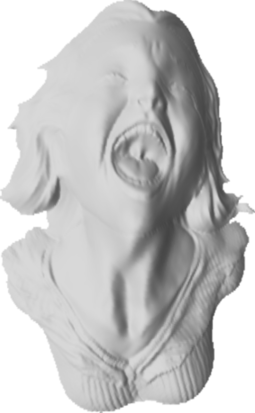}&
    \includegraphics[width=\mywidth]{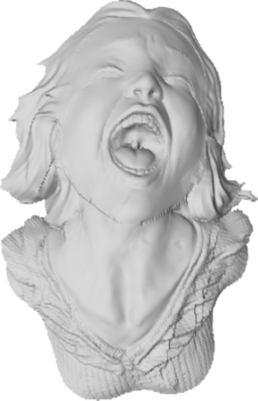}&
    \includegraphics[width=\mywidth]{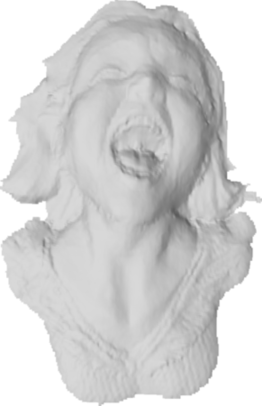}&
    \includegraphics[width=\mywidth]{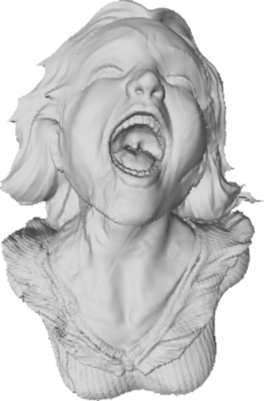}&
    \includegraphics[width=\mywidth]{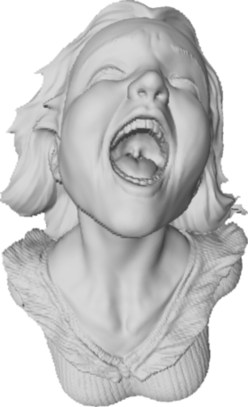}\\
    &MAE:& $21.33^\circ$ & $16.33^\circ$ & $19.70^\circ$ & $\mathbf{9.21^\circ}$ & \\\hline
    \rotatebox{90}{Thai Statue \& White}&
    \includegraphics[width=\mywidth, height=60pt]{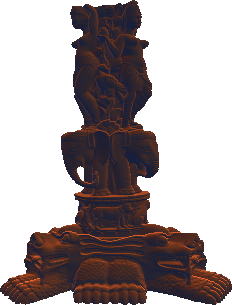}&
    \includegraphics[width=\mywidth]{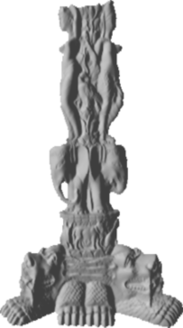}&
    \includegraphics[width=\mywidth]{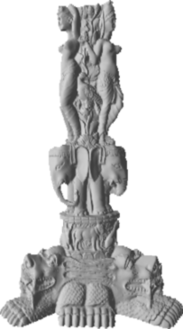}&
    \includegraphics[width=\mywidth]{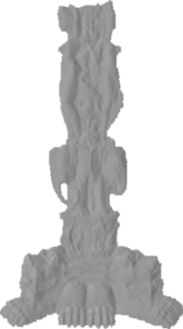}&
    \includegraphics[width=\mywidth]{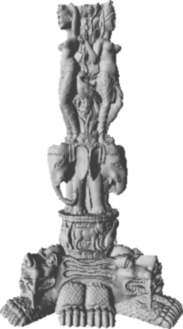}&
    \includegraphics[width=\mywidth]{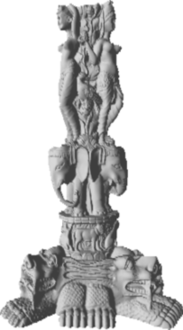}\\
    &MAE:& $28.02^\circ$ & $18.64^\circ$ & $37.31^\circ$ & $\mathbf{9.16^\circ}$ &
  \end{tabular}
  \caption{Results of state-of-the-art approaches and our approach on two out of the 36  synthetic datasets. Numbers show the mean angular error (MAE) in degrees.}
  \label{fig:synthetic_quantitative}
\end{figure}

Having fixed the choice of $(V, \mu, M)$, we can now evaluate our approach against other state-of-the-art methods. We compare our results against those obtained by an uncalibrated photometric stereo approach assuming directional lighting~\cite{Favaro2012}, and another one assuming general (first-order spherical harmonics) illumination yet relying on an input shape prior (e.g., from an RGB-D sensor)~\cite{Peng2017}. As this limiting assumption on the access to a sensor-based depth prior is not always given and to make comparison fair, we input as depth prior to this method the ballooning initialization described in Section~\ref{sec:5.1}. Furthermore, we compare against another uncalibrated photometric stereo work under natural illumination \cite{Mo2018}\footnote{Code associated with~\cite{Favaro2012} and~\cite{Peng2017} can be found online, and the results obtained by~\cite{Mo2018} were provided by the authors.}, which resorts to the equivalent directional lighting instead of spherical harmonics, cf. Section~\ref{sec:ifsha}. Table~\ref{tab:synthetic_quantitative} shows the median and mean MAEs over all 36 datasets (a more detailed table can be found in the supplementary material). On these datasets, it can be seen that our method quantitatively outperforms the current state-of-the-art by a factor of $2$--$3$. This gain is also evaluated qualitatively in Figure~\ref{fig:synthetic_quantitative}, which shows a selection of two results. 

\begin{table}[!h]                                                                                                                  
\centering                                                                                                                       
\begin{tabular}{|c|c|c|c|c|}                                                                                                   
\hline                                                                                                                           
Approach  & \cite{Favaro2012} & \cite{Peng2017} & \cite{Mo2018} & {Ours} \\\hline
Median    & 27.16             & 21.14           & 34.06         & \textbf{9.17} \\\hline
Mean      & 34.15             & 21.18           & 35.53         & \textbf{10.72} \\\hline
\end{tabular}
\caption{Median and mean of the mean angular errors (MAE) over all 36 datasets. The proposed approach overcomes the state-of-the-art by a factor of $2$--$3$.}
\label{tab:synthetic_quantitative}                                                                                        
\end{table}                                                                                                                     

\subsection{Real-World Experiments}\label{seq:rw_exp}
\begin{figure}[!ht]
  \centering
  \newcommand{\mywidth}{0.0766\textwidth} 
  \newcommand{\mywidthy}{0.09\textwidth} 
  \newcolumntype{C}{ >{\centering\arraybackslash} m{0.02\textwidth} }
  \newcolumntype{X}{ >{\centering\arraybackslash} m{\mywidth} }
  \setlength\tabcolsep{3.3pt} 
  \def\arraystretch{1} 
  \begin{tabular}{CXXXXX}
    &$I^i$&\cite{Favaro2012}&\cite{Peng2017}&\cite{Mo2018}& {Ours}\\
    \\[-.4cm]
    \rotatebox{90}{Face 1}&
    \includegraphics[width=\mywidth]{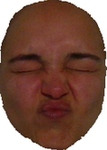}&
    \includegraphics[width=\mywidth]{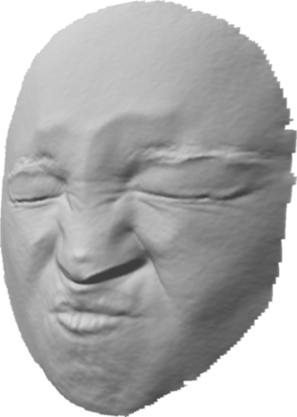}&
    \includegraphics[width=\mywidth]{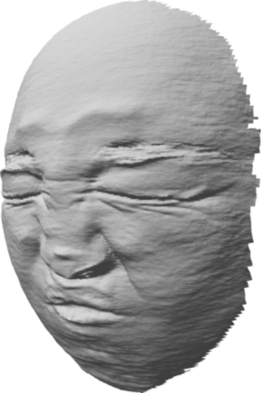}&
    \includegraphics[width=\mywidth]{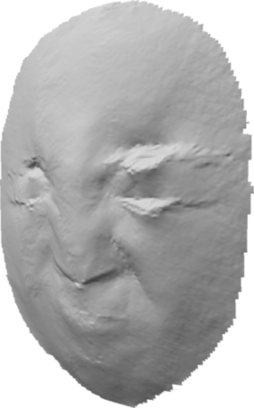}&
    \includegraphics[width=\mywidth]{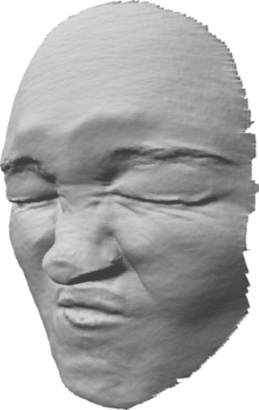}\\
    \rotatebox{90}{Face 2}&
    \includegraphics[width=\mywidth]{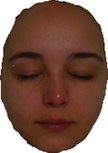}&
    \includegraphics[width=\mywidth]{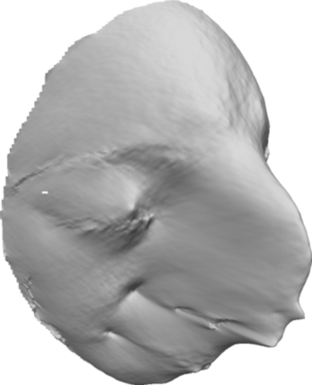}&
    \includegraphics[width=\mywidth]{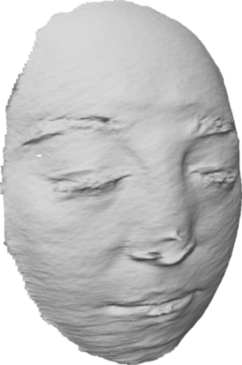}&
    \includegraphics[width=\mywidth]{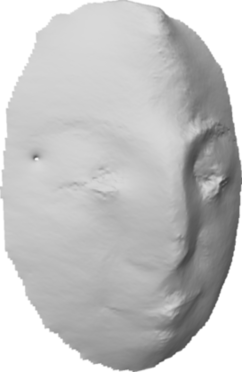}&
    \includegraphics[width=\mywidth]{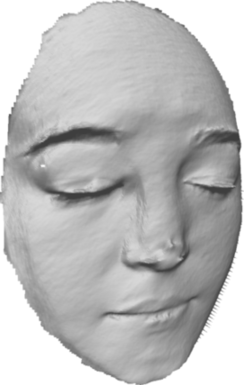}\\
    \rotatebox{90}{Rucksack}&
    \includegraphics[width=\mywidthy]{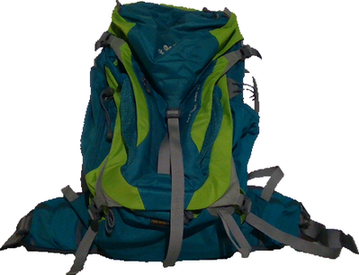}&
    \includegraphics[width=\mywidthy]{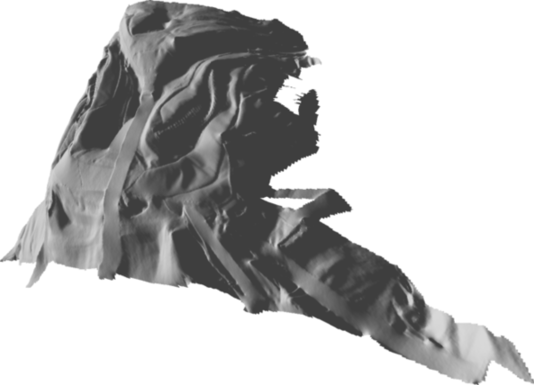}&
    \includegraphics[width=\mywidthy]{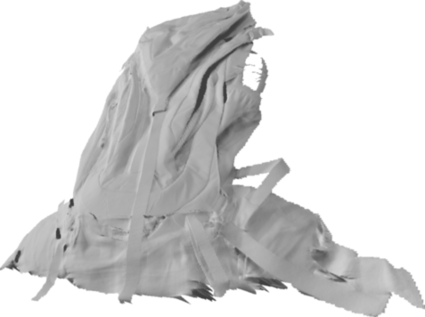}&
    \includegraphics[width=\mywidthy]{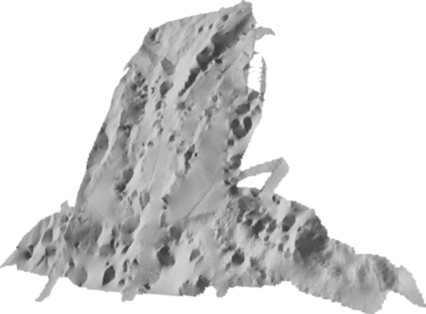}&
    \includegraphics[width=\mywidthy]{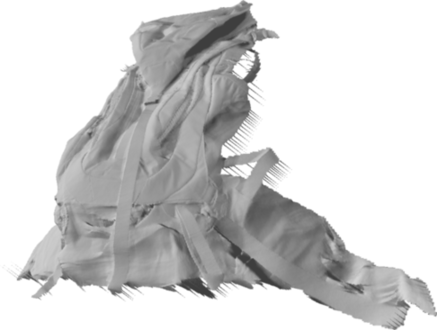}\\
    \rotatebox{90}{Backpack}&
    \includegraphics[width=\mywidth]{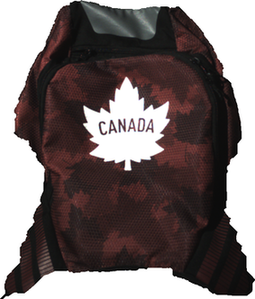}&
    \includegraphics[width=\mywidth]{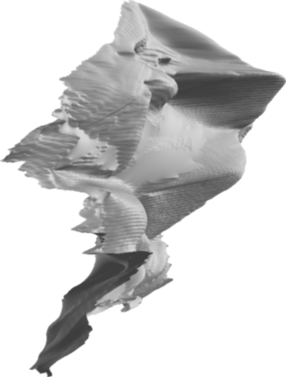}&
    \includegraphics[width=\mywidth]{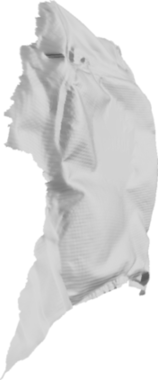}&
    \includegraphics[width=\mywidth]{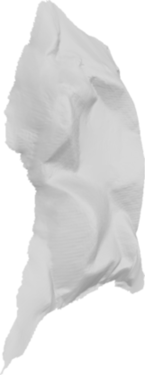}&
    \includegraphics[width=\mywidth]{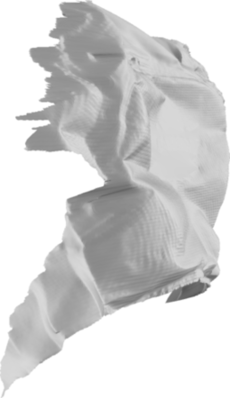}\\
    \rotatebox{90}{Ovenmitt}&
    \includegraphics[width=\mywidth]{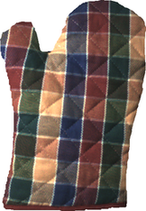}&
    \includegraphics[width=\mywidth]{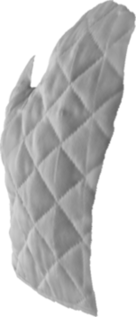}&
    \includegraphics[width=\mywidth]{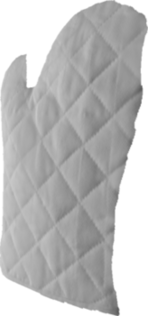}&
    \includegraphics[width=\mywidth]{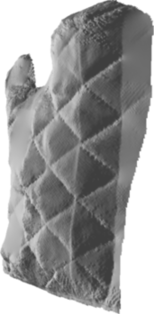}&
    \includegraphics[width=\mywidth]{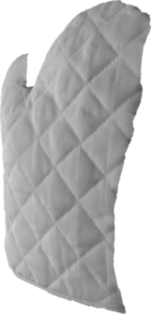}\\
    \rotatebox{90}{Shirt}&
    \includegraphics[width=\mywidth]{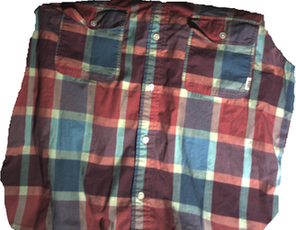}&
    \includegraphics[width=\mywidth]{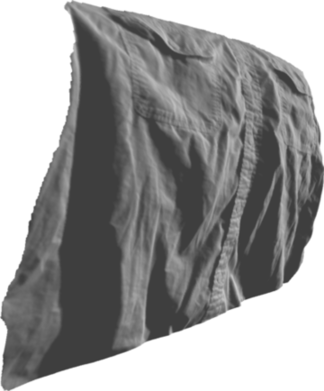}&
    \includegraphics[width=\mywidth]{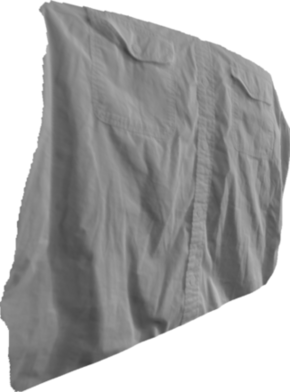}&
    \includegraphics[width=\mywidth]{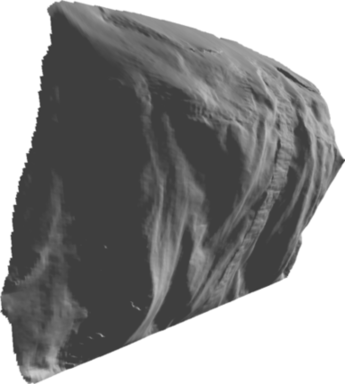}&
    \includegraphics[width=\mywidth]{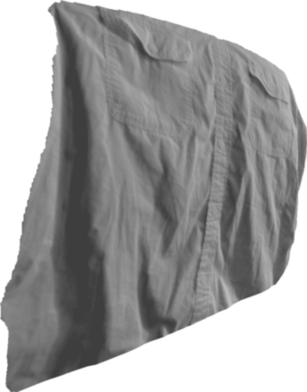}\\
    \rotatebox{90}{Tabletcase}&
    \includegraphics[width=\mywidth]{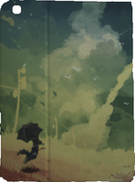}&
    \includegraphics[width=\mywidth]{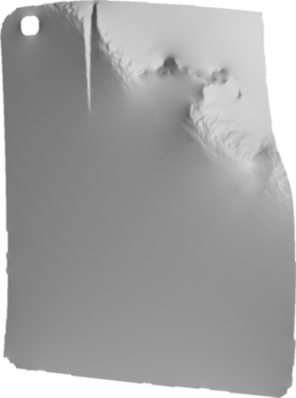}&
    \includegraphics[width=\mywidth]{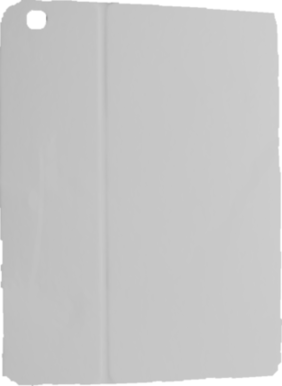}&
    \includegraphics[width=\mywidth]{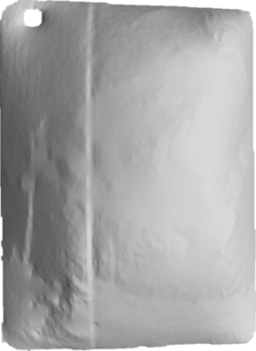}&
    \includegraphics[width=\mywidth]{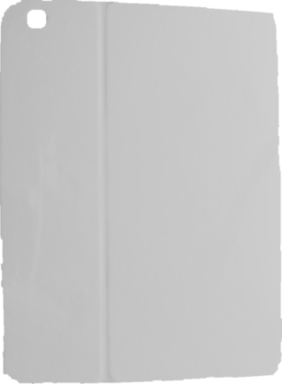}\\
    \rotatebox{90}{Vase}&
    \includegraphics[width=\mywidth]{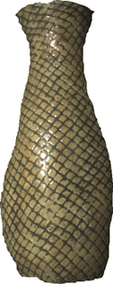}&
    \includegraphics[width=\mywidth]{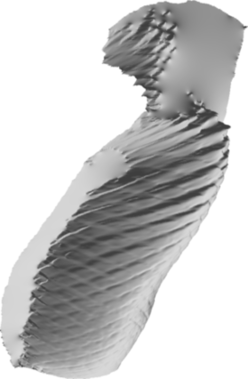}&
    \includegraphics[width=\mywidth]{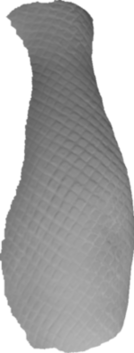}&
    \includegraphics[width=\mywidth]{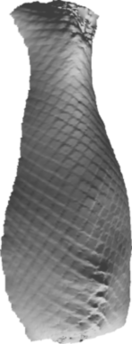}&
    \includegraphics[width=\mywidth]{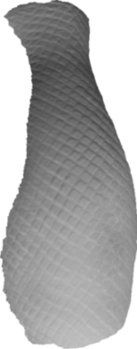}
  \end{tabular}
  \caption{Results of state-of-the-art approaches and our approach on challenging real-world datasets. While the competing approaches fail on some datasets, our approach consistently yields satisfactory results.}
\label{fig:realworld_qualitative}
\end{figure}

For real-world data we use the publicly available dataset of~\cite{Haefner2018b}. It offers eight challenging real-world datasets of objects with complex geometry and albedo captured under daylight and a freely moving LED, along with intrinsics matrix $K$ and masks $\Omega$. Results are presented in Figure~\ref{fig:realworld_qualitative}. Despite relying on a directional lighting model, the approach of~\cite{Favaro2012} produces reasonable results on some datasets (Face1, Ovenmitt or Shirt), but it fails on others. As~\cite{Peng2017} assumes a reliable prior on depth in order to perform a photometric refinement, this approach is biased towards its initialization and thus, only when the depth prior is very close to the objects' rough shape (Ovenmitt, Shirt, Tabletcase, Vase) a meaningful geometry is recovered. The approach of~\cite{Mo2018} estimates a possibly non-integrable normal field only, and it can be seen that after integration the depth map might not be satisfactory. As our approach optimizes over depth directly, such issues are not apparent and we are able to recover fine-scale geometric details throughout all tests.

\section{Conclusion}\label{sec:7}
We proposed a variational approach to uncalibrated photometric stereo (PS) under general lighting. Assuming a perspective camera setup, our method jointly estimates shape, reflectance and lighting in a robust manner. The possible non-integrability of normals is bypassed by the direct estimation of the underlying depth map, and robustness is ensured by resorting to Cauchy's M-estimator and Huber-TV albedo regularization. Although the problem is nonconvex and thus numerically challenging and initialization-dependent, we tackled it efficiently through a tailored lagged block coordinate descent algorithm and ballooning-based depth initialization. Over a series of evaluations on synthetic and real data, we demonstrated that our method outperforms existing methods in terms of MAE by a factor of $2$--$3$ and provides highly detailed reconstructions even in challenging real-world settings. 

In future research, a more automated balloon-like depth initialization is desirable.
Exploring the theoretical foundations (uniqueness of a solution) of differential perspective uncalibrated PS under spherical harmonic lighting and analyzing the convergence properties of the proposed numerical scheme constitute two other promising perspectives.

\clearpage
\appendix
\section{Further Details on Synthetic Experiments}
\begin{figure*}[!ht]
  \centering
  \newcommand{\mywidth}{0.196\textwidth} 
  \newcolumntype{X}{ >{\centering\arraybackslash} m{\mywidth} }
  \setlength\tabcolsep{1pt} 
  \def\arraystretch{1} 
  \begin{tabular}{XXXXX}
    \includegraphics[width=\mywidth]{lighting/alexs_apt_2k}&
    \includegraphics[width=\mywidth]{lighting/arches_e_pinetree_3k}&
    \includegraphics[width=\mywidth]{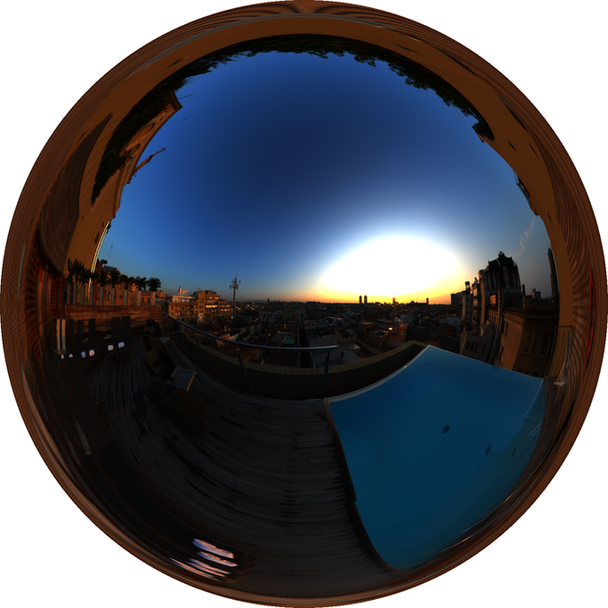}&
    \includegraphics[width=\mywidth]{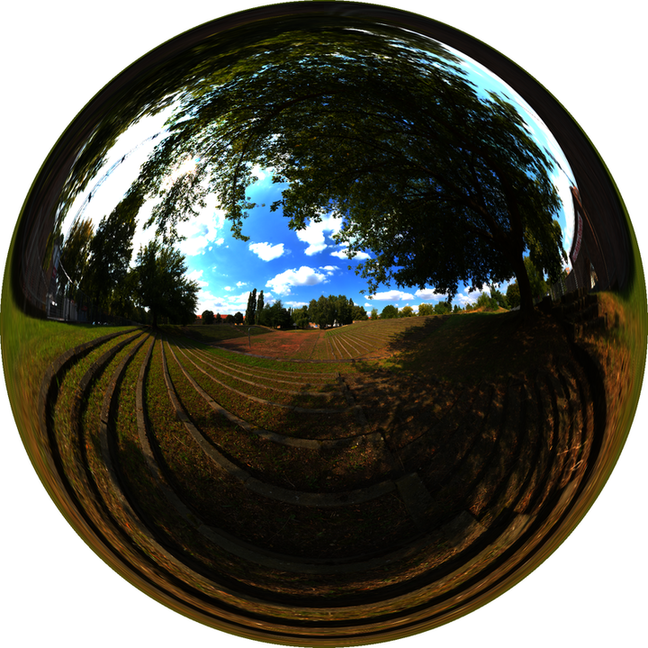}&
    \includegraphics[width=\mywidth]{lighting/brooklyn_bridge_planks_2k}\\
    \includegraphics[width=\mywidth]{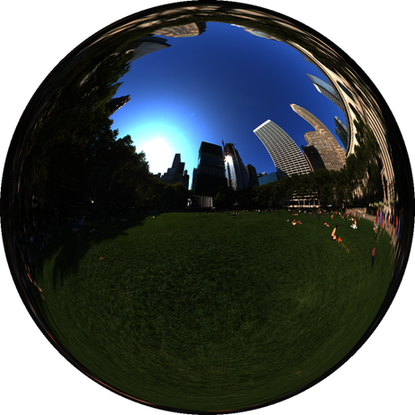}&
    \includegraphics[width=\mywidth]{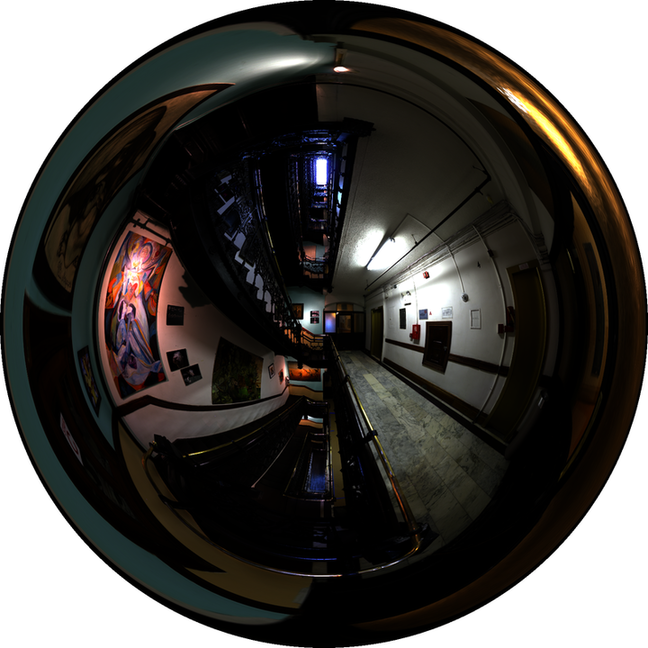}&
    \includegraphics[width=\mywidth]{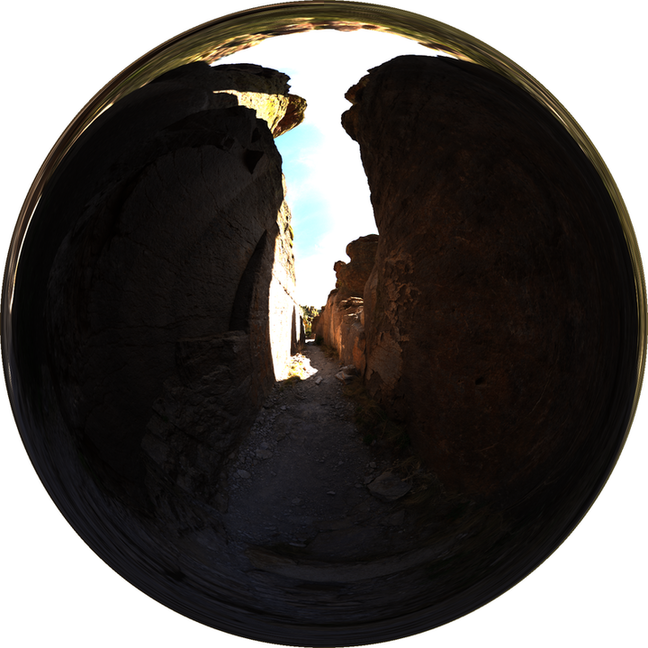}&
    \includegraphics[width=\mywidth]{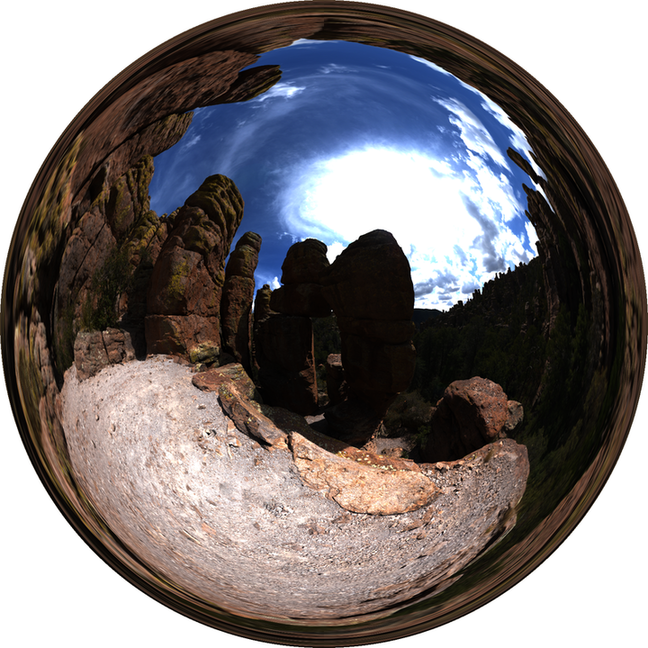}&
    \includegraphics[width=\mywidth]{lighting/circus_backstage_3k}\\
    \includegraphics[width=\mywidth]{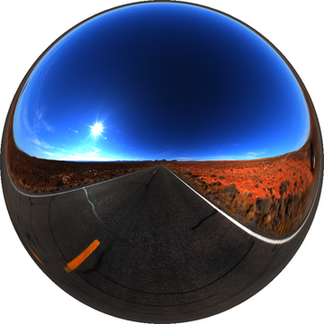}&
    \includegraphics[width=\mywidth]{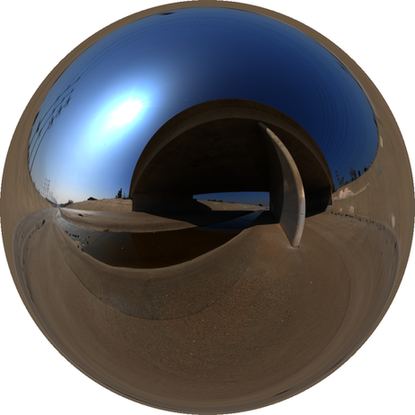}&
    \includegraphics[width=\mywidth]{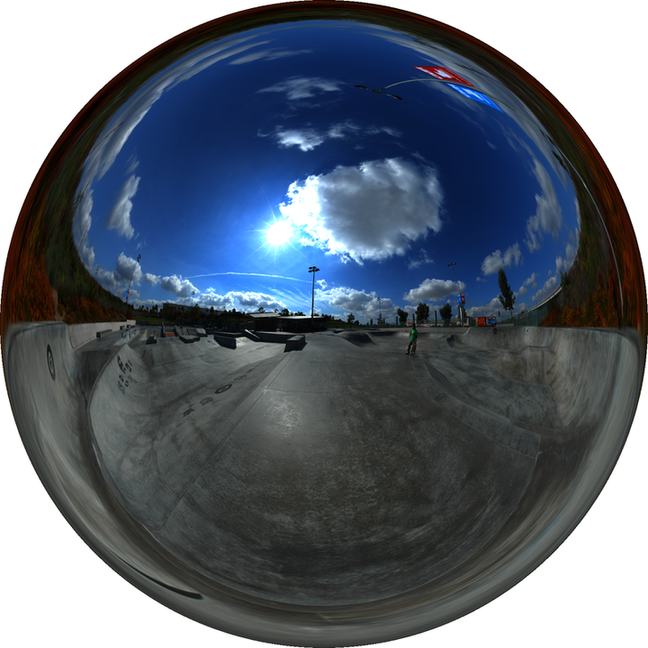}&
    \includegraphics[width=\mywidth]{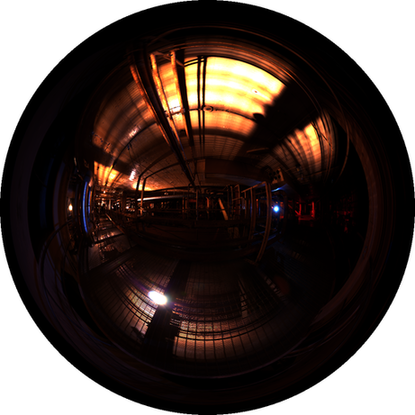}&
    \includegraphics[width=\mywidth]{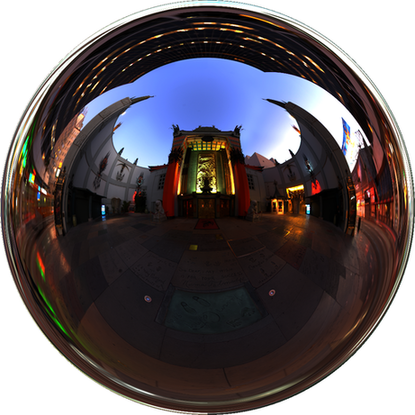}\\
    \includegraphics[width=\mywidth]{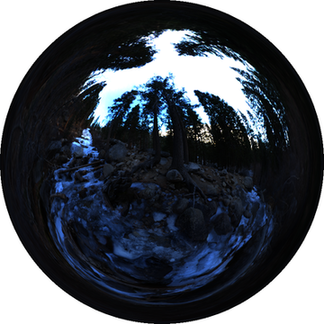}&
    \includegraphics[width=\mywidth]{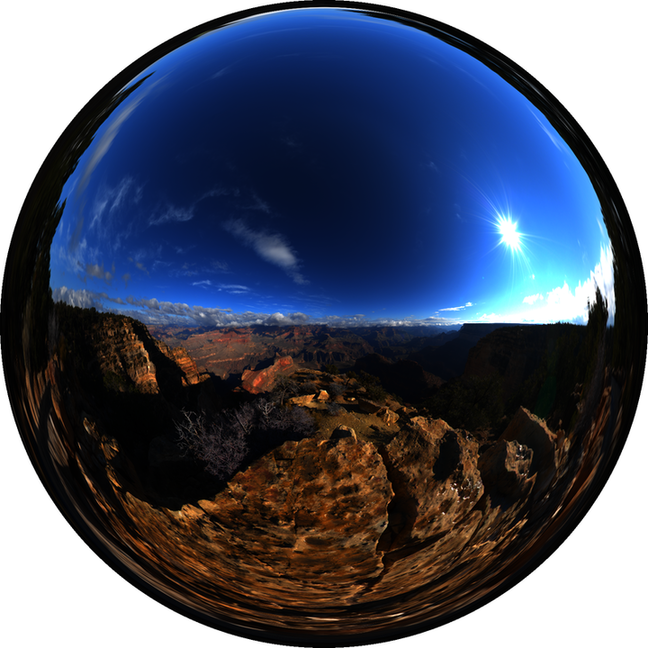}&
    \includegraphics[width=\mywidth]{lighting/14-hamarikyu_bridge_b_3k}&
    \includegraphics[width=\mywidth]{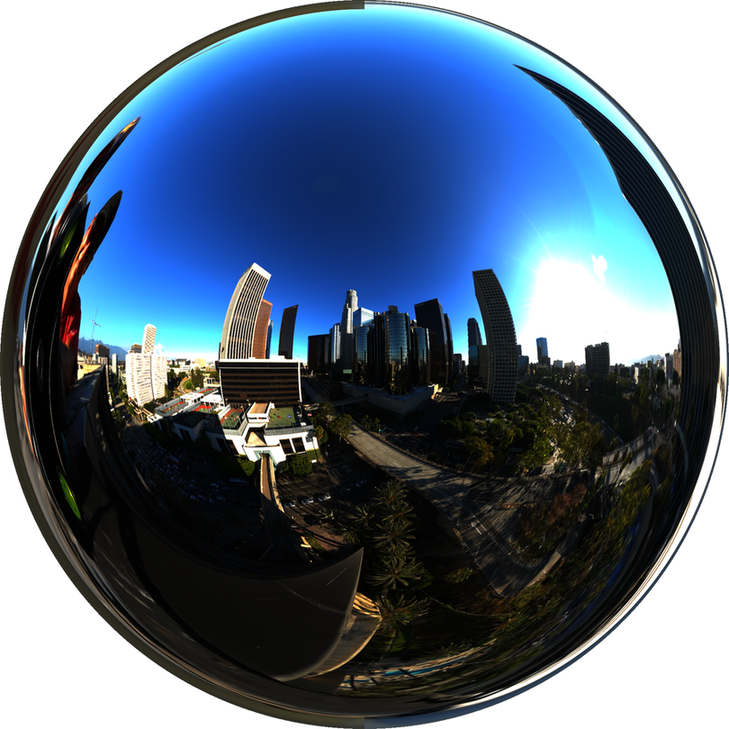}&
    \includegraphics[width=\mywidth]{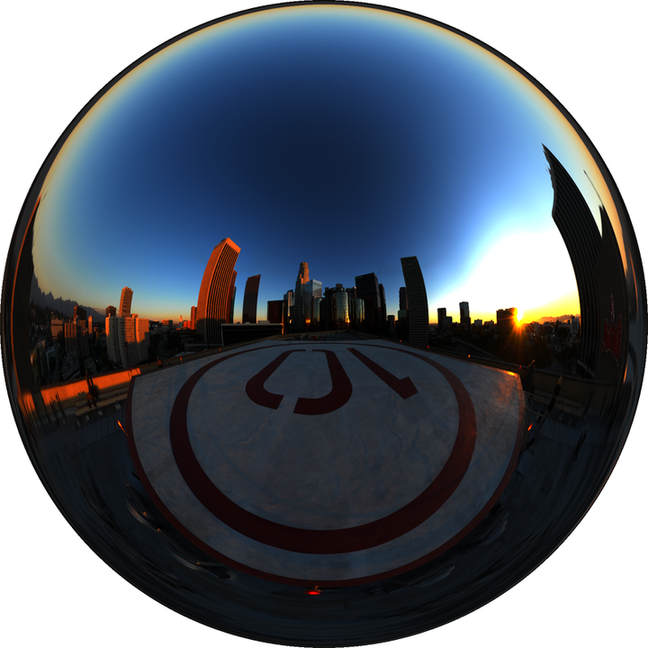}\\
    \includegraphics[width=\mywidth]{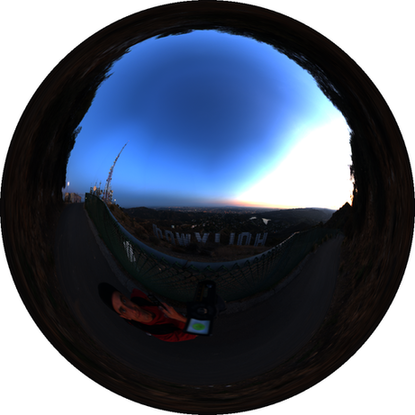}&
    \includegraphics[width=\mywidth]{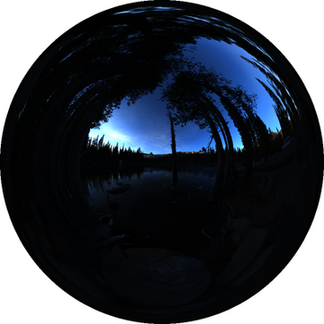}&
    \includegraphics[width=\mywidth]{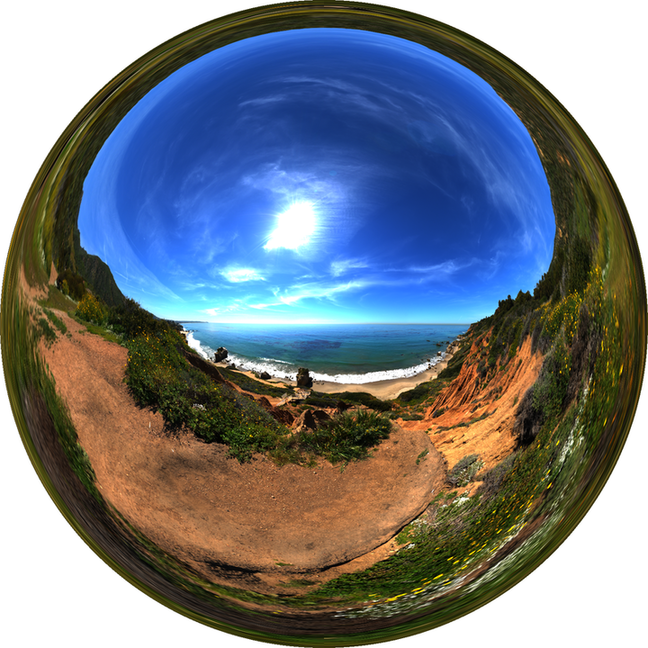}&
    \includegraphics[width=\mywidth]{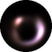}&
    \includegraphics[width=\mywidth]{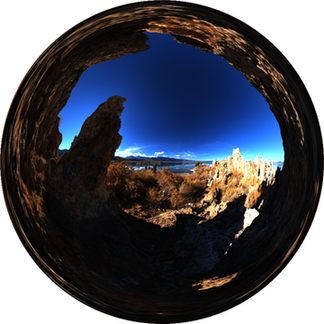}\\
   \end{tabular}
  \caption{All environment maps $\ell^i$ ($360^\circ$ view) used throughout the synthetic evaluation.}
  \label{fig:supp_envmaps}
\end{figure*}

To provide further insights on the synthetic experiments (in Section 6.1), we visualize the environment lightings $\ell^i$, $i=1\dots25$, used to render each image. Figure~\ref{fig:supp_envmaps} shows all $25$ environment maps\footnote{All environment maps were downloaded from \url{http://www.hdrlabs.com/sibl/archive.html}}.

\begin{figure*}[!ht]
  \centering
  \newcommand{\mywidth}{0.196\textwidth} 
  \newcolumntype{X}{ >{\centering\arraybackslash} m{\mywidth} }
  \setlength\tabcolsep{1pt} 
  \def\arraystretch{1} 
  \begin{tabular}{XXXXX}
    \includegraphics[width=\mywidth]{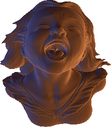}&
    \includegraphics[width=\mywidth]{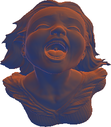}&
    \includegraphics[width=\mywidth]{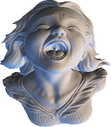}&
    \includegraphics[width=\mywidth]{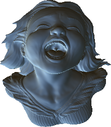}&
    \includegraphics[width=\mywidth]{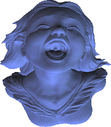}\\
     \includegraphics[width=\mywidth]{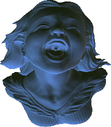}&
    \includegraphics[width=\mywidth]{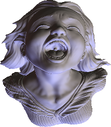}&
    \includegraphics[width=\mywidth]{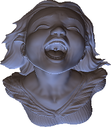}&
    \includegraphics[width=\mywidth]{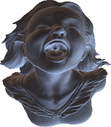}&
    \includegraphics[width=\mywidth]{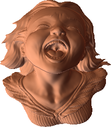}\\
     \includegraphics[width=\mywidth]{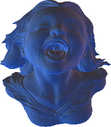}&
    \includegraphics[width=\mywidth]{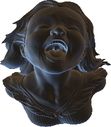}&
    \includegraphics[width=\mywidth]{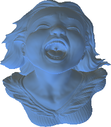}&
    \includegraphics[width=\mywidth]{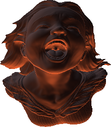}&
    \includegraphics[width=\mywidth]{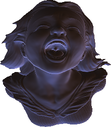}\\
     \includegraphics[width=\mywidth]{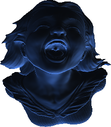}&
    \includegraphics[width=\mywidth]{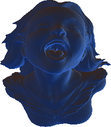}&
    \includegraphics[width=\mywidth]{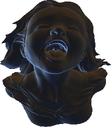}&
    \includegraphics[width=\mywidth]{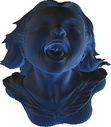}&
    \includegraphics[width=\mywidth]{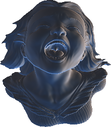}\\
     \includegraphics[width=\mywidth]{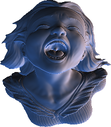}&
    \includegraphics[width=\mywidth]{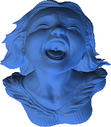}&
    \includegraphics[width=\mywidth]{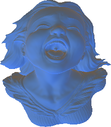}&
    \includegraphics[width=\mywidth]{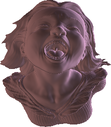}&
    \includegraphics[width=\mywidth]{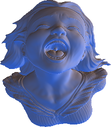}\\
   \end{tabular}
  \caption{Illustration of the input data. The Joyful Yell dataset with White albedo to show the impact of the different environment maps used throughout the synthetic experimental validation.}
  \label{fig:illustration_input}
\end{figure*}

The impact of each incident lighting $\ell^i$, $i=1\dots25$, is illustrated in Figure~\ref{fig:illustration_input} showing the Joyful Yell with a White ($\rho\equiv1$) albedo. Thus, color changes in the images are caused by lighting only, as depicted in model (1) and (7) in the main paper.\\

Table~\ref{tab:supp_synthetic_quantitative_detailed} shows the mean angular error (MAE) of each dataset on the state-of-the-art approaches~\cite{Favaro2012,Mo2018,Peng2017} and our proposed methodology. It can be seen that our approach consistently overcomes \cite{Favaro2012,Mo2018,Peng2017} by a factor of $2$--$3$. Only the Pattern albedo seems to bias the resulting depth negatively, yet even in this case our approach estimates the geometry more faithfully than the current state-of-the-art.\\
\begin{table*}                                                                                                                   
\centering                                                                                         
\setlength\tabcolsep{12.8pt}                              
\begin{tabular}{|c|c||c|c|c|c|c|c|}                                                                                                   
\hline                                                                                                                           
\multicolumn{2}{|c||}{Dataset} & 
\multirow{3}{*}{\cite{Favaro2012}} & 
\multirow{3}{*}{\cite{Peng2017}} & 
\multirow{3}{*}{\cite{Mo2018}} & 
\multicolumn{3}{c|}{Our approach with different initializations}  \\\cline{1-2} \cline{6-8}
\multirow{2}{*}{Shape} & 
\multicolumn{1}{c||}{\multirow{2}{*}{Albedo}} & & & & \multirow{2}{*}{Hemisphere} &
\multirow{2}{*}{Using \cite{Mo2018}} & 
Minimal surface\\
&&&&&&&(Sec. 5.1)   
\\ \hline
\multirow{9}{*}{\rotatebox{90}{Armadillo}}
& Bars & 26.22 & 27.84 & 36.91 & 79.54 & 20.08 & \textbf{16.78}\\                                                                     
& Constant & 25.84 & 26.64 & 36.87 & 83.01 & 18.81 & \textbf{13.97}\\                                                                 
& Ebsd & 25.34 & 26.88 & 27.80 & 82.53 & 15.99 & \textbf{14.26}\\                                                                     
& Hippie & 28.21 & 27.30 & 25.82 & 79.12 & \textbf{12.56} & 14.52\\                                                                   
& Lena & 27.07 & 27.33 & 28.36 & 84.24 & 17.79 & \textbf{14.78}\\                                                                     
& Pattern & 45.87 & 26.82 & 24.01 & 82.59 & 19.39 & \textbf{19.06}\\                                                                  
& Rectcircle & 26.97 & 26.71 & 36.23 & 80.68 & 19.64 & \textbf{14.06}\\                                                               
& Voronoi & 25.62 & 26.91 & 50.70 & 79.65 & 55.29& \textbf{14.07}\\                                                                  
& White & 26.19 & 26.64 & 52.04 & 83.04 & 56.74 & \textbf{14.13}\\                                                                    
\hline                                                                                                                           
\multirow{9}{*}{\rotatebox{90}{Joyful Yell}} 
& Bars & 21.84 & 16.26 & 31.80 & 21.21 & 28.82 & \textbf{8.69}\\                                                                     
& Constant & 23.95 & 14.93 & 33.47 & 16.85 & 29.31 & \textbf{5.96}\\                                                                 
& Ebsd & 26.08 & 15.63 & 15.91 & 17.63 & 7.49 & \textbf{7.28}\\                                                                     
& Hippie & 28.67 & 16.23 & 22.96  & 17.68 & \textbf{7.47} & 7.49\\                                                                   
& Lena & 21.33 & 16.33 & 19.70 & 20.11 & 13.16 & \textbf{9.21}\\                                                                     
& Pattern & 26.07 & 18.76 & 26.67 & 18.76 & 21.03 & \textbf{16.97}\\                                                                 
& Rectcircle & 35.27 & 15.19 & 52.41 & 16.27 & 61.77 & \textbf{7.34}\\                                                               
& Voronoi & 22.27 & 16.42 & 45.74 & 18.62 & 54.78 & \textbf{6.57}\\                                                                  
& White & 27.12 & 14.32 & 33.06 & 17.70 & 28.99 & \textbf{6.20}\\                                                                    
\hline                                                                                                                           
\multirow{9}{*}{\rotatebox{90}{Lucy}} 
& Bars & 49.13 & 21.90 & 36.51 & 40.55 & 26.15 & \textbf{8.16}\\                                                                           
& Constant & 54.98 & 19.89 & 36.57 & 41.00 & 25.74 & \textbf{8.71}\\                                                                       
& Ebsd & 62.33 & 20.81 & 23.56 & 40.80 & 13.36  & \textbf{9.61}\\                                                                           
& Hippie & 58.61 & 21.29 & 32.38 & 39.93 & 8.10 & \textbf{7.87}\\                                                                         
& Lena & 64.01 & 22.24 & 30.93 & 40.16 & 19.14 & \textbf{9.56}\\                                                                           
& Pattern & 48.83 & 22.25 & 32.68 & 40.11 & 20.56 & \textbf{17.78}\\                                                                       
& Rectcircle & 24.68 & 20.99 & 43.13 & 41.17 & 10.01  & \textbf{8.98}\\                                                                     
& Voronoi & 61.53 & 22.10 & 48.14 & 40.39 & 71.32 & \textbf{7.59}\\                                                                        
& White & 64.43 & 19.33 & 44.76 & 41.54 & 72.45 & \textbf{8.76}\\                                                                          
\hline                                                                                                                           
\multirow{9}{*}{\rotatebox{90}{Thai Statue}} 
& Bars & 25.53 & 21.91 & 66.17 & 78.72 & 8.94 & \textbf{8.55} \\                                                                     
& Constant & 27.20 & 18.91 & 38.47 & 81.14 & 24.26 & \textbf{9.58}\\                                                                 
& Ebsd & 27.85 & 20.22 & 34.11 & 79.58 & 19.23 & \textbf{9.47}\\                                                                     
& Hippie & 21.91 & 21.86 & 30.62 & 77.27 & 12.78 & \textbf{8.83}\\                                                                   
& Lena & 33.53 & 19.66 & 34.00 & 79.43 & 19.55 & \textbf{9.19}\\                                                                     
& Pattern & 26.77 & 22.06 & 28.81 & 83.92 & 16.69  & \textbf{15.27}\\                                                                 
& Rectcircle & 29.36 & 19.92 & 43.86 & 81.88 & 79.88 & \textbf{8.84}\\                                                               
& Voronoi & 30.65 & 21.56 & 36.58 & 78.92 & 25.21 & \textbf{8.69}\\                                                                  
& White & 28.02 & 18.64 & 37.31 & 81.54 & 24.94 & \textbf{9.16}\\                                                                    
\hline\hline
\multicolumn{2}{|c||}{Median} & 27.16 & 21.14 & 34.06 & 59.41 & 19.86 & \textbf{9.17}\\                                                                       
\hline                                                                                                                           
\multicolumn{2}{|c||}{Mean} & 34.15 & 21.18 & 35.53 & 55.20 & 27.43 & \textbf{10.72}\\                                                                          
\hline                                                                                                                           
\end{tabular}                                                                                                                    
\caption{Quantitative comparison between our method and other state-of-the-art methods on challenging synthetic datasets. The last three columns refer to the results with different initializations for our approach.}
\label{tab:supp_synthetic_quantitative_detailed}                                                                                         
\end{table*}                                                                                                                     

\begin{figure*}[!ht]
  \centering
  \newcommand{\mywidth}{0.15\textwidth} 
  \newcommand{\myheight}{0.24\textheight} 
  \newcolumntype{C}{ >{\centering\arraybackslash} m{0.02\textwidth} }
  \newcolumntype{X}{ >{\centering\arraybackslash} m{\mywidth} }
  \setlength\tabcolsep{3pt} 
  \def\arraystretch{1} 
  \begin{tabular}{CXXXXXX}
    &$I^i$&\cite{Favaro2012}&\cite{Peng2017}&\cite{Mo2018}&{Ours}& GT\\\hline
    \rotatebox{90}{Armadillo \& Constant}&
    \includegraphics[width=\mywidth, height=3cm]{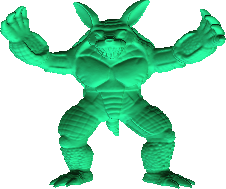}&
    \includegraphics[width=\mywidth]{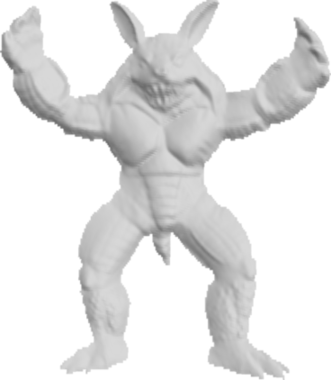}&
    \includegraphics[width=\mywidth]{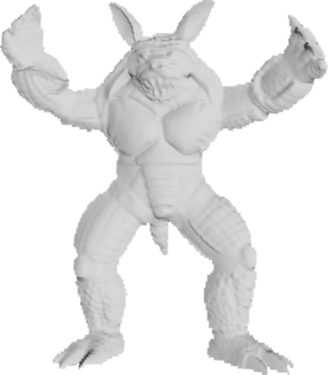}&
    \includegraphics[width=\mywidth]{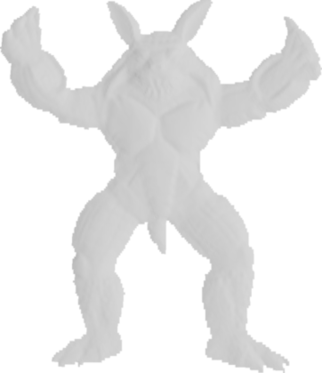}&
    \includegraphics[width=\mywidth]{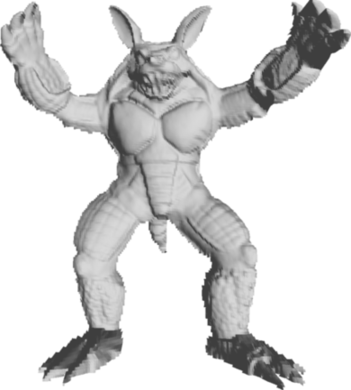}&
    \includegraphics[width=\mywidth]{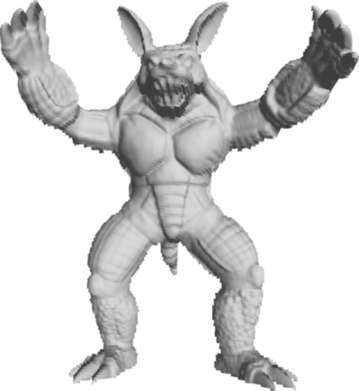}\\
    &MAE:& $25.84^\circ$ & $26.64^\circ$ & $27.80^\circ$ & $\mathbf{13.97^\circ}$ & \\\hline
    \rotatebox{90}{Lucy \& Hippie}&
    \includegraphics[width=\mywidth, height=5.5cm]{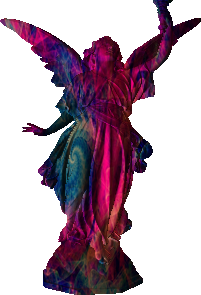}&
    \includegraphics[width=\mywidth, height=\myheight]{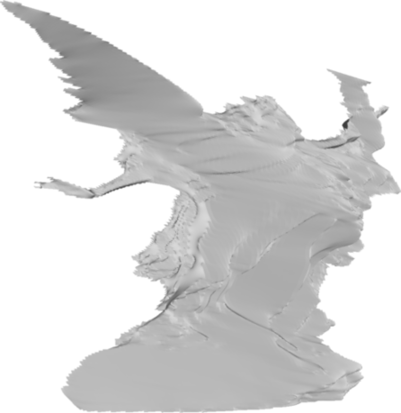}&
    \includegraphics[width=\mywidth, height=\myheight]{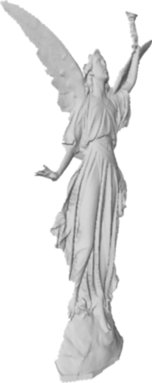}&
    \includegraphics[width=\mywidth, height=\myheight]{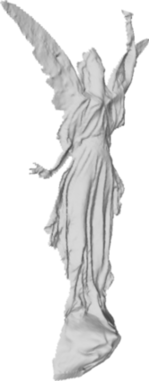}&
    \includegraphics[width=\mywidth]{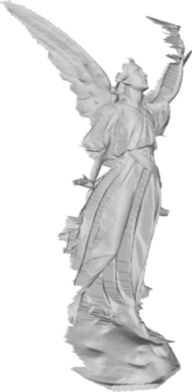}&
    \includegraphics[width=\mywidth, height=\myheight]{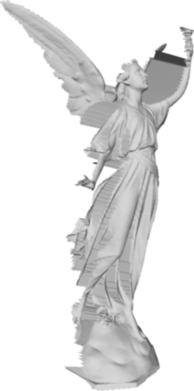}\\
    &MAE:& $58.61^\circ$ & $21.29^\circ$ & $32.38^\circ$ & $\mathbf{7.87^\circ}$ &
  \end{tabular}
  \caption{Results of state-of-the-art approaches and our approach on two out of the 36  synthetic datasets. Numbers show the mean angular error (MAE) in degrees.}
  \label{fig:supp_synthetic_qualitative}
\end{figure*}

Two more qualitative results on synthetic data are shown in Figure~\ref{fig:supp_synthetic_qualitative}. While \cite{Favaro2012} gives more meaningful results on Armadillo with Constant albedo, depth deteriorates strongly on Lucy with Hippie albedo. Methods of \cite{Mo2018,Peng2017} both result in rather flattened shapes (cf.\ Lucy). Most accurate results are achieved using the proposed method where fine geometric details, as well as non flattened depth estimates are shown.\\

\begin{figure*}[!ht]
  \centering
  \newcommand{\mywidth}{0.185\textwidth} 
  \newcolumntype{X}{>{\centering\arraybackslash}  m{\mywidth} }
  \newcolumntype{C}{>{\centering\arraybackslash}  m{0.02\textwidth} }
  \setlength\tabcolsep{2.5pt} 
  \def\arraystretch{1} 
  \begin{tabular}{CXXXXX}
  & $I^i$ & Albedo GT & Albedo Ours & Lighting GT & Lighting Ours\\
    \rotatebox{90}{Armadillo \& Constant}&
    \includegraphics[width=\mywidth]{armadillo_constant_rgb}&
    \includegraphics[width=\mywidth]{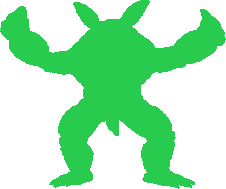}&
    \includegraphics[width=\mywidth]{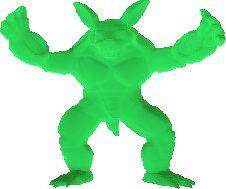}&
    \includegraphics[width=\mywidth]{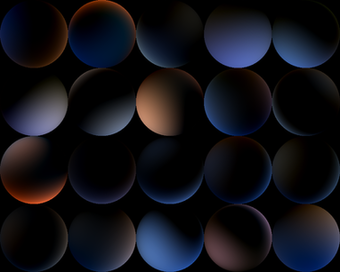}&
    \includegraphics[width=\mywidth]{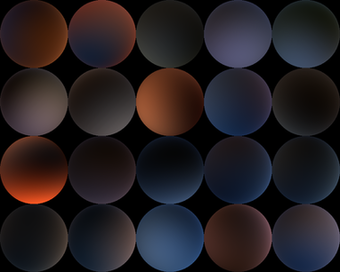}\\
    \rotatebox{90}{Joyful Yell \& Lena}&
    \includegraphics[width=\mywidth]{joyfulyell_lena_rgb}&
    \includegraphics[width=\mywidth]{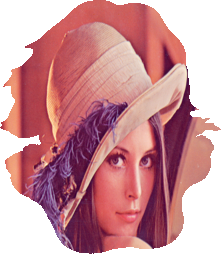}&
    \includegraphics[width=\mywidth]{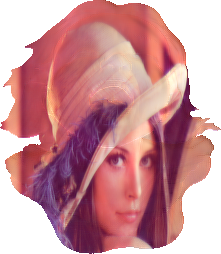}&
    \includegraphics[width=\mywidth]{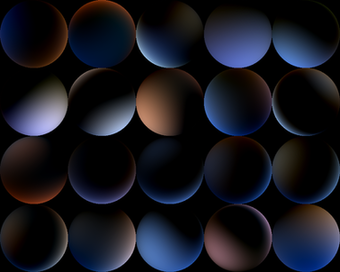}&
    \includegraphics[width=\mywidth]{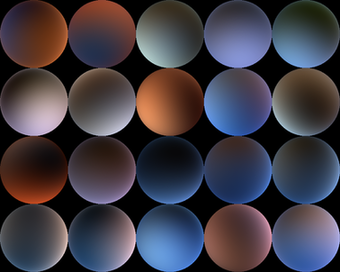}\\
    \rotatebox{90}{Joyful Yell \& White}&
    \includegraphics[width=\mywidth]{joyfulyell_white_rgb_1}&
    \includegraphics[width=\mywidth]{albedos/white}&
    \includegraphics[width=\mywidth]{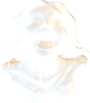}&
    \includegraphics[width=\mywidth]{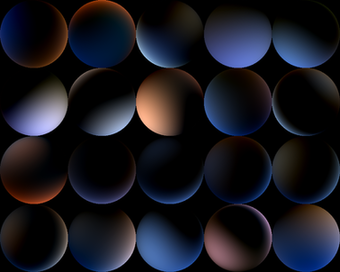}&
    \includegraphics[width=\mywidth]{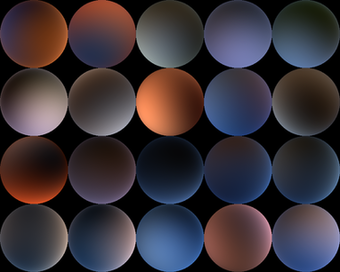}\\
    \rotatebox{90}{Lucy \& Hippie}&
    \includegraphics[width=\mywidth]{lucy_hippie_rgb}&
    \includegraphics[width=\mywidth]{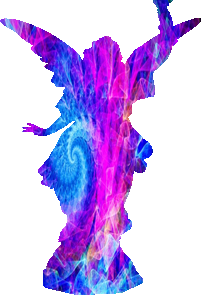}&
    \includegraphics[width=\mywidth]{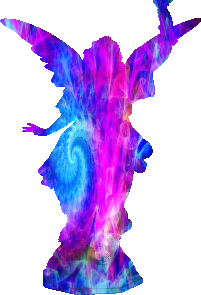}&
    \includegraphics[width=\mywidth]{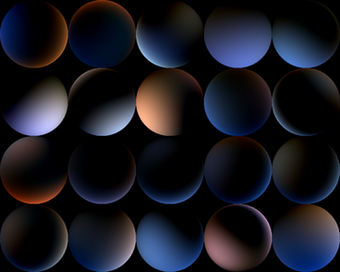}&
    \includegraphics[width=\mywidth]{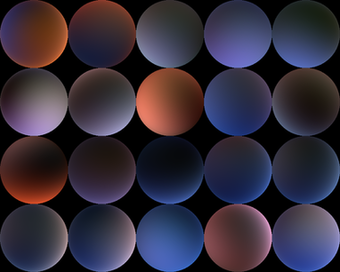}\\
    \rotatebox{90}{Thai Statue \& White}&
    \includegraphics[width=\mywidth]{thaistatue_white_rgb}&
    \includegraphics[width=\mywidth]{albedos/white}&
    \includegraphics[width=\mywidth]{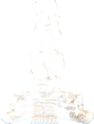}&
    \includegraphics[width=\mywidth]{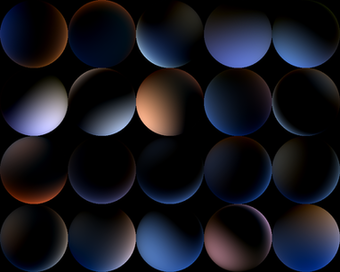}&
    \includegraphics[width=\mywidth]{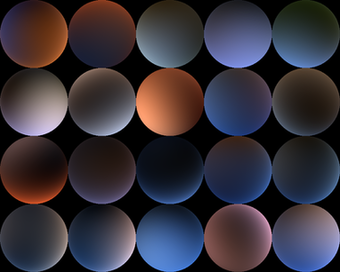}
  \end{tabular}
  \caption{Our estimated albedos and lighting next to the ground truth. Lighting estimates show less shadowed areas and seem brighter compared to ground truth, yet this does not seem to affect reflectance and geometry estimation much, cf.\ Figure 7 in main paper and Figure~\ref{fig:supp_synthetic_qualitative} in the supplementary material. The estimated albedos are satisfactory, although some shading information is slightly visible.}
\label{fig:supp_synthetic_lighting_albedo}
\end{figure*}

Additional to the depth results, Figure~\ref{fig:supp_synthetic_lighting_albedo} shows estimated lightings and albedos along with the ground truths. Although lighting estimates show less shadowed areas and seem brighter compared to ground truths, this does not seem to affect reflectance estimations much. The estimated albedos are satisfactory, although some shading information is slightly visible.\\

\begin{figure*}[!ht]
  \centering
  \newcommand{\mywidth}{0.14\textwidth} 
  \newcolumntype{X}{>{\centering\arraybackslash}  m{\mywidth} }
  \newcolumntype{C}{>{\centering\arraybackslash}  m{0.02\textwidth} }
  \setlength\tabcolsep{4.2pt} 
  \def\arraystretch{1} 
  \begin{tabular}{CXXXXXX}
    & Initialization by hemisphere & Final result by hemisphere & Initialization by \cite{Mo2018} &  Final result by \cite{Mo2018}  & Our final result & GT \\
    \rotatebox{90}{Armadillo \& Constant}&
    \includegraphics[height=3.2cm, width=\mywidth]{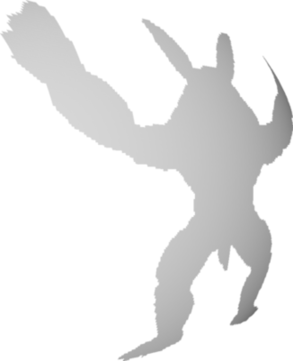}&
    \includegraphics[width=\mywidth]{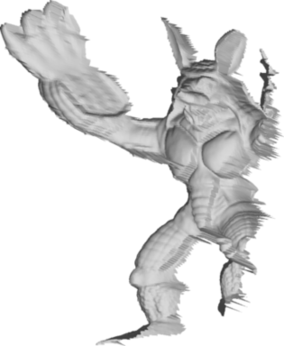}&
    \includegraphics[height=3.2cm, width=\mywidth]{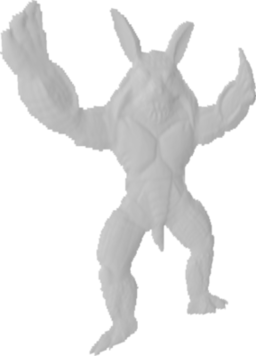}&
    \includegraphics[width=\mywidth]{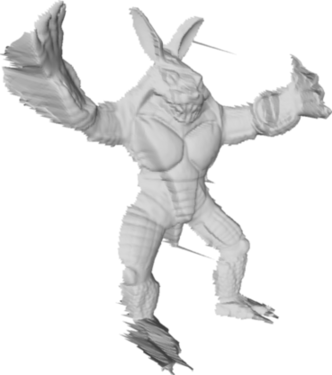}&
    \includegraphics[width=\mywidth]{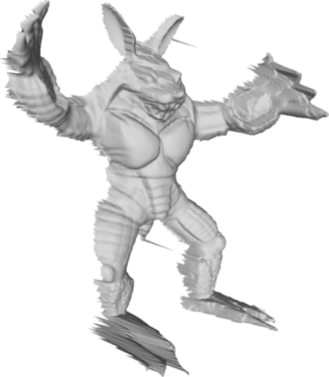}&
    \includegraphics[height=3.2cm,width=\mywidth]{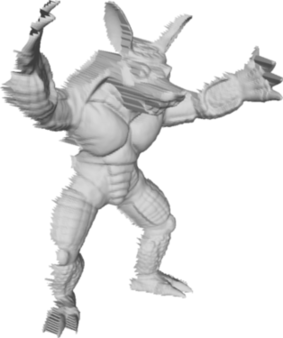}\\
    MAE:& & $83.01^\circ$ & & $18.81^\circ$ & $\mathbf{13.97^\circ}$ &\\
    \rotatebox{90}{Joyful Yell \& Lena}&
    \includegraphics[height=3.7cm, width=\mywidth]{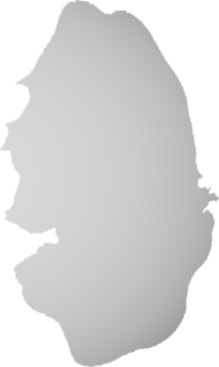}&
    \includegraphics[height=3.7cm,width=\mywidth]{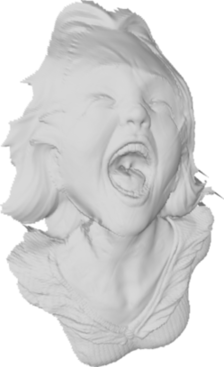}&
    \includegraphics[height=3.7cm, width=\mywidth]{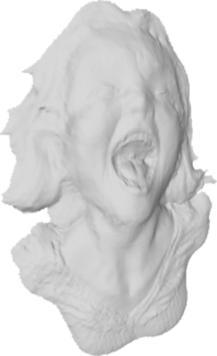}&
    \includegraphics[height=3.7cm, width=\mywidth]{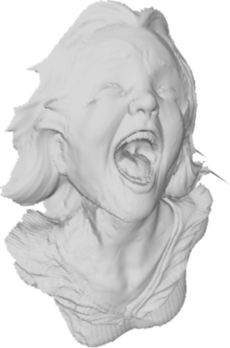}&
    \includegraphics[height=3.7cm, width=\mywidth]{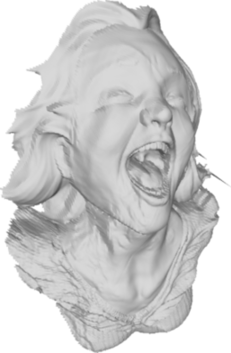}&
    \includegraphics[width=\mywidth]{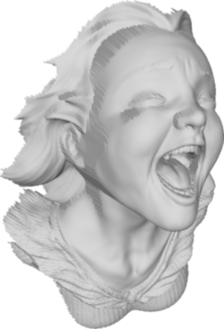}\\
    MAE:& &$20.11^\circ$ && $13.16^\circ$ & $\mathbf{9.21^\circ}$ &\\
    \rotatebox{90}{Joyful Yell \& White}&
    \includegraphics[height=3.7cm, width=\mywidth]{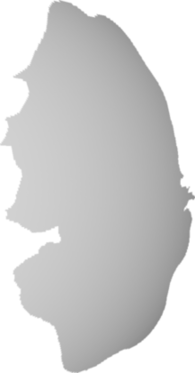}&
    \includegraphics[height=3.7cm, width=\mywidth]{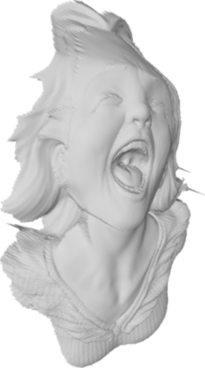}&
    \includegraphics[height=3.7cm, width=\mywidth]{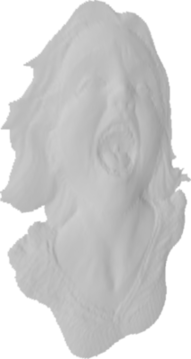}&
    \includegraphics[height=3.7cm, width=\mywidth]{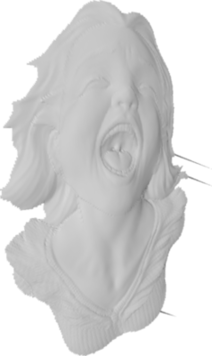}&
    \includegraphics[height=3.7cm, width=\mywidth]{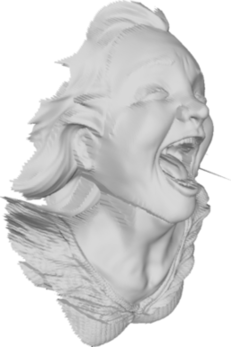}&
    \includegraphics[width=\mywidth]{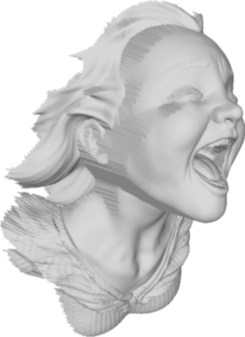}\\
    MAE:& &$17.70^\circ$ && $28.99^\circ$ & $\mathbf{6.20^\circ}$&\\
    \rotatebox{90}{Lucy \& Hippie}&
    \includegraphics[height=4.5cm,width=\mywidth]{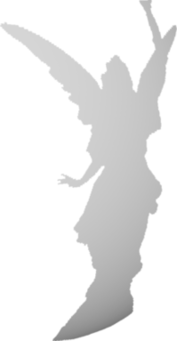}&
    \includegraphics[height=4.5cm,width=\mywidth]{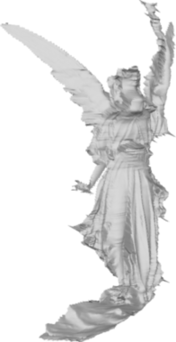}&
    \includegraphics[height=4.5cm,width=\mywidth]{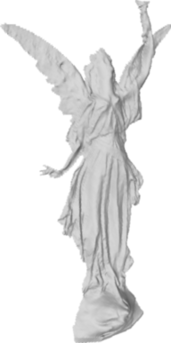}&
    \includegraphics[height=4.6cm, width=\mywidth]{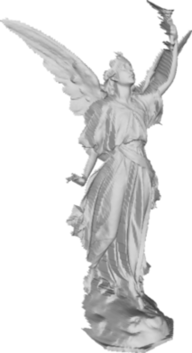}&
    \includegraphics[height=4.6cm, width=\mywidth]{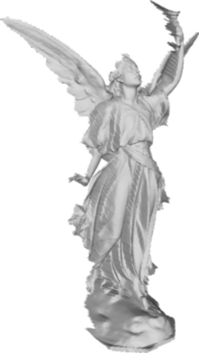}&
    \includegraphics[height=4.6cm,width=\mywidth]{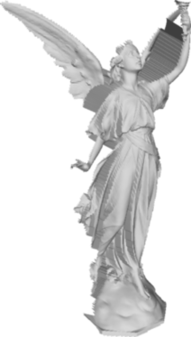}\\
    MAE:& &$39.93^\circ$ && $8.10^\circ$ & $\mathbf{7.87^\circ}$&\\
    \rotatebox{90}{Thai Statue \& White}&
    \includegraphics[height=3.8cm, width=\mywidth]{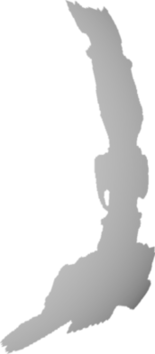}&
    \includegraphics[height=3.8cm, width=\mywidth]{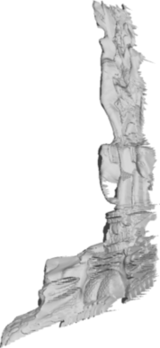}&
    \includegraphics[height=3.8cm, width=\mywidth]{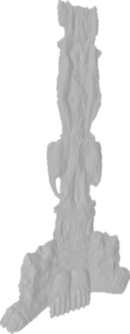}&
    \includegraphics[height=3.8cm, width=\mywidth]{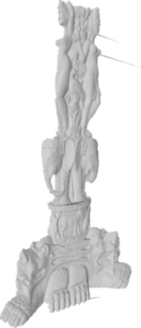}&
    \includegraphics[height=3.8cm, width=\mywidth]{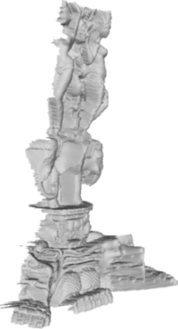}&
    \includegraphics[height=3.8cm, width=\mywidth]{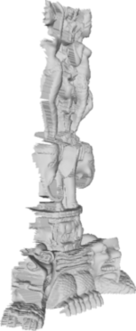}\\
    MAE:& &$81.54^\circ$ && $24.94^\circ$ & $\mathbf{9.16^\circ}$&\\
  \end{tabular}
  \caption{Our results compared those from two different initializations of our algorithm. Numbers show the mean angular error (MAE) in degrees. Though the initialization by \cite{Mo2018} achieves comparable result to ground truth on ``Lucy \& Hippie'' dataset, its performance is not stable across different datasets.}
\label{fig:supp_synthetic_different_init}
\end{figure*}

The initialization is indeed crucial for the whole algorithm. Here, we show two different non-trivial initializations for our algorithm in Table~\ref{tab:supp_synthetic_quantitative_detailed}: 1) Hemisphere, we first compute the circumscribed sphere for the 3D points of ground truth. The projection of each point onto this sphere is considered as initialization; 2) Initialization by \cite{Mo2018}, we simply refine the result from \cite{Mo2018} by our algorithm. In Figure~\ref{fig:supp_synthetic_different_init}, we show visualized results. In certain special cases, the initialization from \cite{Mo2018} is slightly better. However, our minimal surface strategy is stable for all cases, and our algorithm improves the results from \cite{Mo2018}) in most cases.

\section{Further Details on Real-World Results}
\begin{figure*}[!ht]
  \centering
  \newcommand{\mywidth}{0.20\textwidth} 
  \newcommand{\lightwidth}{0.24\textwidth}
  \newcommand{\mycolwidth}{0.2\textwidth} 
  \newcommand{\myheight}{0.11\textheight} 
  \newcolumntype{X}{>{\centering\arraybackslash}  m{\lightwidth} }
  \newcolumntype{C}{>{\centering\arraybackslash}  m{0.02\textwidth} }
  \setlength\tabcolsep{17pt} 
  \def\arraystretch{1} 
  \begin{tabular}{CXXX}
  	& Novel viewpoint & Estimated albedo & Estimated lighting \\
    \rotatebox{90}{Face 1}&
    \includegraphics[width=\mywidth]{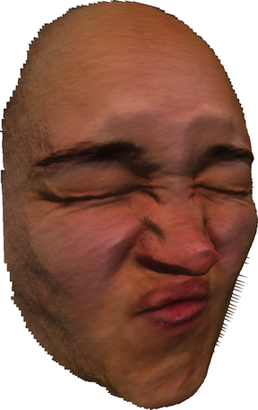}&
    \includegraphics[width=\mywidth]{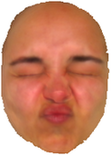}&
    \includegraphics[width=\lightwidth]{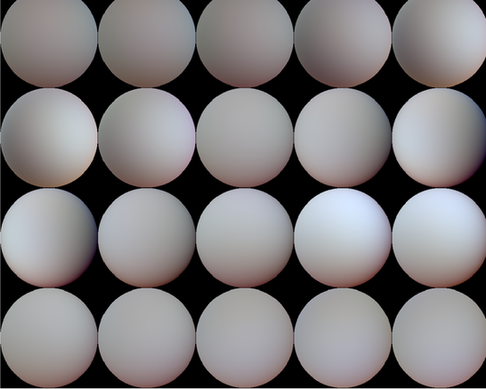}\\
    \rotatebox{90}{Face 2}&
    \includegraphics[width=\mywidth]{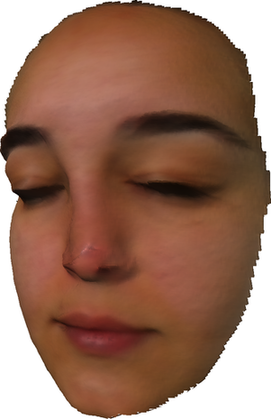}&
    \includegraphics[width=\mywidth]{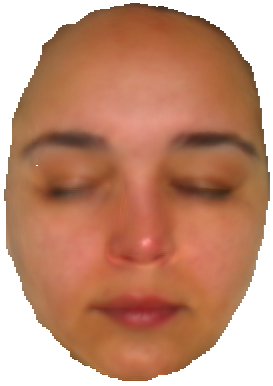}&
    \includegraphics[width=\lightwidth]{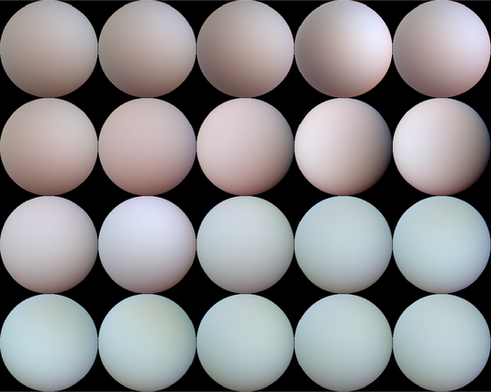}\\
    \rotatebox{90}{Rucksack}&
    \includegraphics[width=\mycolwidth]{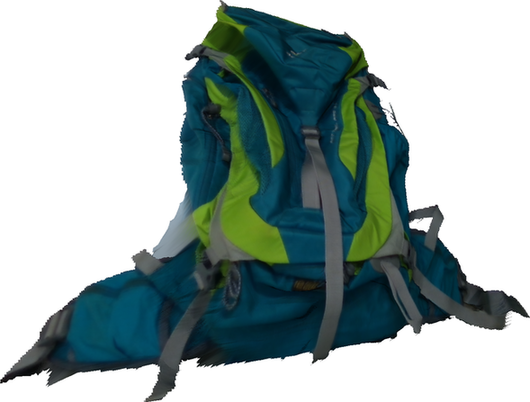}&
    \includegraphics[width=\mywidth]{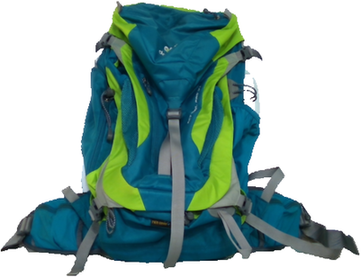}&
    \includegraphics[width=\lightwidth]{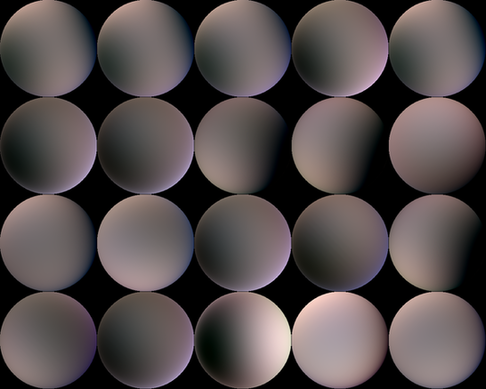}\\
    \rotatebox{90}{Backpack}&
    \includegraphics[width=\mywidth]{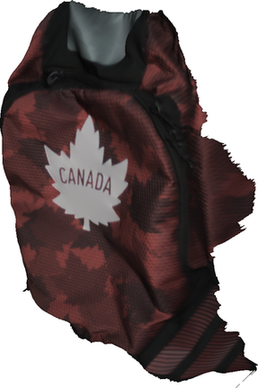}&
    \includegraphics[width=\mywidth]{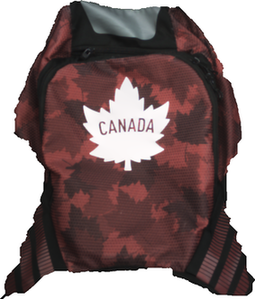}&
    \includegraphics[width=\lightwidth]{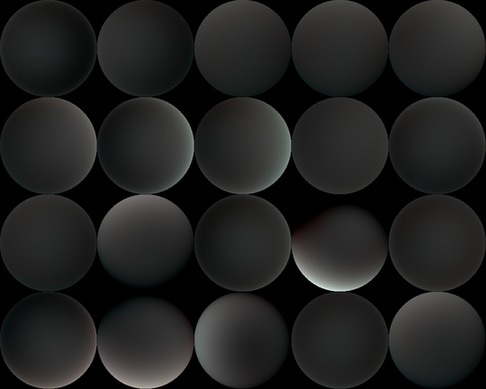}\\
  \end{tabular}
  \caption{Real-world results: (left) estimated albedos mapped onto estimated surfaces rendered under a novel viewpoint, (middle) estimated albedos, (right) estimated lightings for all $M=20$ input images.}
\label{fig:supp_realworld_qualitative1}
\end{figure*}

\begin{figure*}[!ht]
	\centering
\newcommand{\mywidth}{0.15\textwidth} 
\newcommand{\lightwidth}{0.24\textwidth}
\newcommand{\mycolwidth}{0.2\textwidth} 
\newcommand{\myheight}{0.11\textheight} 
\newcolumntype{X}{>{\centering\arraybackslash}  m{\lightwidth} }
\newcolumntype{C}{>{\centering\arraybackslash}  m{0.02\textwidth} }

	\setlength\tabcolsep{17pt} 
	\def\arraystretch{1} 
	\begin{tabular}{CXXX}
		& Novel viewpoint & Estimated albedo & Estimated lighting \\
		\rotatebox{90}{Ovenmitt}&    \includegraphics[width=\mywidth]{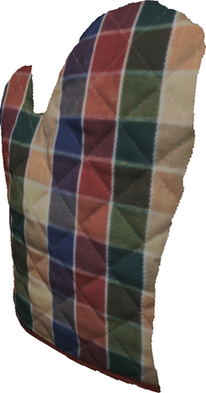}&
		\includegraphics[width=\mywidth]{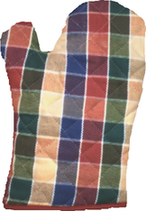}&
		\includegraphics[width=\lightwidth]{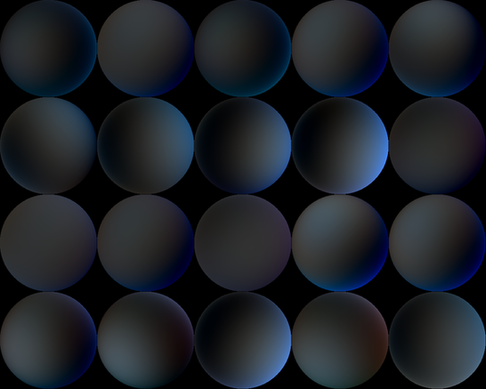}\\
		\rotatebox{90}{Shirt}&
		\includegraphics[width=0.15\textwidth]{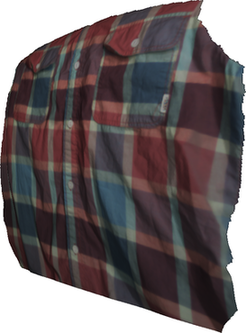}&
		\includegraphics[width=\lightwidth]{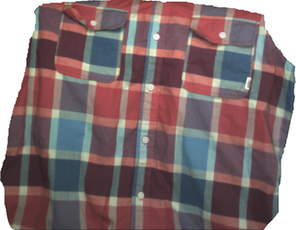}&
		\includegraphics[width=\lightwidth]{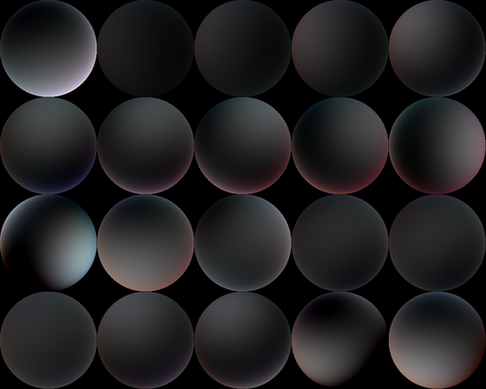}\\
		\rotatebox{90}{Tabletcase}&
		\includegraphics[width=\mywidth]{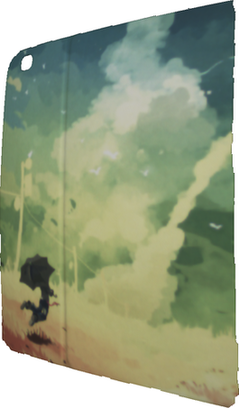}&
		\includegraphics[width=\mywidth]{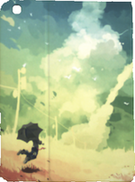}&
		\includegraphics[width=\lightwidth]{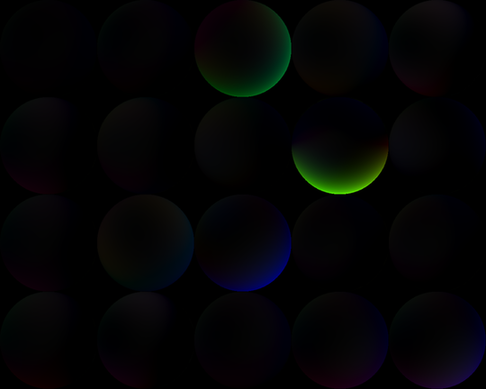}\\
		\rotatebox{90}{Vase}&
		\includegraphics[width=\mywidth]{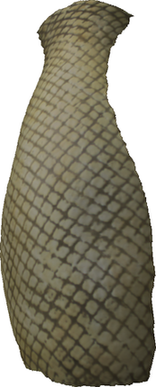}&
		\includegraphics[width=\mywidth]{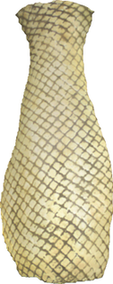}&
		\includegraphics[width=\lightwidth]{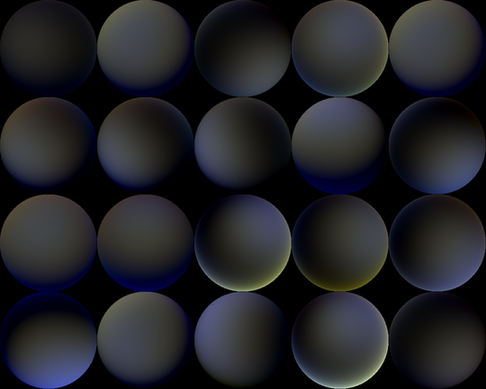}\\
	\end{tabular}
	\caption{More real-world results: (left) estimated albedos mapped onto estimated surfaces rendered under a novel viewpoint, (middle) estimated albedos, (right) estimated lightings for all $M=20$ input images.}
	\label{fig:supp_realworld_qualitative2}
\end{figure*}

Supplementary to the real-world experiments (in Section 6.2), Figures~\ref{fig:supp_realworld_qualitative1} and \ref{fig:supp_realworld_qualitative2} show alternative viewpoints of the real-world results. The estimated albedos, which are mapped onto the surfaces, appear satisfactory. Correspondingly, we also show the estimated albedos and lightings. In view of the multiplicative ambiguity between lightings and albedos, all visualized albedos are normalized to have maximum value 1.\\

\clearpage \clearpage
{\small
	\bibliographystyle{ieee_fullname}
	\bibliography{biblio}
}

\end{document}